\documentclass[smallextended, twocolumn]{svjour3}     
\smartqed

\usepackage{amsmath} 
\usepackage{amssymb} 
\usepackage{amsfonts}
\usepackage{mathtools}
\usepackage{times}  
\usepackage{balance}
\usepackage{graphicx} 
\usepackage{blindtext}
\usepackage[hidelinks]{hyperref}
\usepackage{multirow}
\usepackage{tabularx}
\usepackage{booktabs}
\usepackage{subfig}
\usepackage{float}
\usepackage{color}
\usepackage{cite}
\usepackage[flushleft]{threeparttable}
\usepackage[export]{adjustbox}
\usepackage[linesnumbered, ruled]{algorithm2e}

\definecolor{axesRed}{RGB}{214,39,40}
\definecolor{axesGreen}{RGB}{44,160,44}
\definecolor{axesBlue}{RGB}{31,119,180}
\definecolor{myGold}{RGB}{255,192,0}

\begin{document}

\title{ Incorporating Learnt Local and Global Embeddings into Monocular Visual SLAM }
\author{Huaiyang Huang \and  Haoyang Ye \and Yuxiang Sun \and Lujia Wang \and Ming Liu* }

\institute{
	This work was supported by the National Natural Science Foundation of China, under grant No. U1713211, Collaborative Research Fund by Research Grants Council Hong Kong, under Project No. C4063-18G, and HKUST-SJTU Joint Research Collaboration Fund, under project SJTU20EG03, awarded to Prof. Ming Liu. Ming Liu is the corresponding author.\\
	\\
	\and Huaiyang Huang \at
	\email{hhuangat@connect.ust.hk}           
	\and
	Haoyang Ye \at
	\email{hy.ye@connect.ust.hk}           
	\and
	Lujia Wang \at
	\email{eewanglj@ust.hk}           
	\and
	Ming Liu (\textit{corresponding author})\at
	\email{eelium@ust.hk}           
	\and
	Department of Electronic and Computer Engineering, \\
	The Hong Kong University of Science and Technology, \\
	Hong Kong, SAR, China \\
	\\
	\and Yuxiang Sun \at
	\email{yx.sun@polyu.edu.hk, sun.yuxiang@outlook.com}           
	\and
	Department of Mechanical Engineering, \\
	The Hong Kong Polytechnic University, Hung Hom\\
	Hong Kong, SAR, China \\
}

\newcommand\Tstrut{\rule{0pt}{2.6ex}}         
\newcommand\Bstrut{\rule[-0.9ex]{0pt}{0pt}}   

\maketitle

\begin{abstract}
	Traditional approaches for Visual Simultaneous Localization and Mapping (VSLAM) rely on low-level vision information for state estimation, such as handcrafted local features or the image gradient.
	While significant progress has been made through this track,
	under more challenging configuration for monocular VSLAM, e.g., varying illumination, the performance of state-of-the-art systems generally degrades.
	As a consequence, robustness and accuracy for monocular VSLAM are still widely concerned.
	This paper presents a monocular VSLAM system that fully exploits learnt features for better state estimation.
	The proposed system leverages both learnt local features and global embeddings at different modules of the system: direct camera pose estimation, inter-frame feature association, and loop closure detection.
	With a probabilistic explanation of keypoint prediction, we formulate the camera pose tracking in a direct manner and parameterize local features with uncertainty taken into account.
	To alleviate the quantization effect, we adapt the mapping module to generate 3D landmarks better to guarantee the system's robustness.
	Detecting temporal loop closure via deep global embeddings further improves the robustness and accuracy of the proposed system.
	The proposed system is extensively evaluated on public datasets (Tsukuba, EuRoC, and KITTI), and compared against the state-of-the-art methods.
	The competitive performance of camera pose estimation confirms the effectiveness of our method.
\end{abstract}


\section{INTRODUCTION}
\label{sec:intro}

Visual Simultaneous Localization and Mapping (VSLAM) is a fundamental building block for robotic state estimation \cite{cadena2016past}, providing critical information for high-level applications.
With the recent progress in this field, VSLAM has been well developed for various real-world applications, e.g., visual-based navigation for autonomous vehicles,
ranging from path planning to decision making.
Moreover, VSLAM can model the environment and build a corresponding map apart from ego-motion estimation. This capability can further benefit the robots in scene understanding and object recognition.

Despite the impressive progress under moderate scenarios,
it is noticeable that diverse challenges \cite{SUN2018}, for instance, adverse illumination conditions \cite{park2017illumination}, could cause failure to most of the current VLSAM systems.
Therefore, the robustness and accuracy under challenging situations remain a problem demanding further investigation.
Emerging advances in deep learning (DL) provide an alternative perspective in resolving the aforementioned challenges.
For examples,
leveraging learning-based depth prediction \cite{mahjourian2018unsupervised}, object recognition \cite{liu2016ssd}, or high-level semantics \cite{long2015fully} benefits the visual state estimation in several aspects, varying from local reconstruction \cite{loo2019cnn}, scale recovery \cite{tateno2017cnn, yang2019cubeslam} to dynamic environment handling \cite{runz2018maskfusion}.
Especially in recent years, the exploration of learning local feature extraction and description is prevailing \cite{balntas2017hpatches}.
Following a \textit{detect-and-describe} scheme to predict the repeatability and description within one forward pass, several works \cite{detone2018superpoint,dusmanu2019d2,tang2019gcnv2} demonstrate a superior efficiency against the traditional \textit{detect-then-describe} pipelines, e.g., SIFT \cite{lowe2004distinctive}.
The larger receptive fields from convolutional neural networks (CNNs) and metric learning techniques further contribute to the competitive performance of learning-based methods in both detection and description.
These evolutions pave the way for the introduction of deep features into VSLAM systems.
While some existing works generally recover pose in an end-to-end manner \cite{wang2017deepvo,li2018undeepvo}, limiting the generalization ability for different scenarios,
we consider a more practical approach to utilize the deep feature.

Built on top of ORB-SLAM2 \cite{mur2017orb}, we propose a monocular SLAM system powered by learnt local and global embeddings.
Leveraging the repeatability probability predicted from a CNN,
the initial camera pose can be tracked robustly in a direct manner.
Pre-triangulated landmarks can then be associated with local observations efficiently, which establishes correspondences for the pose refinement.
We discuss the parameterization of local features to facilitate different modules from pose estimation to mapping.
To achieve global consistency, we further leverage deep global embeddings to detect temporal loop closure.

\begin{figure}[t!]
	\centering
		\includegraphics[width=0.5\textwidth]{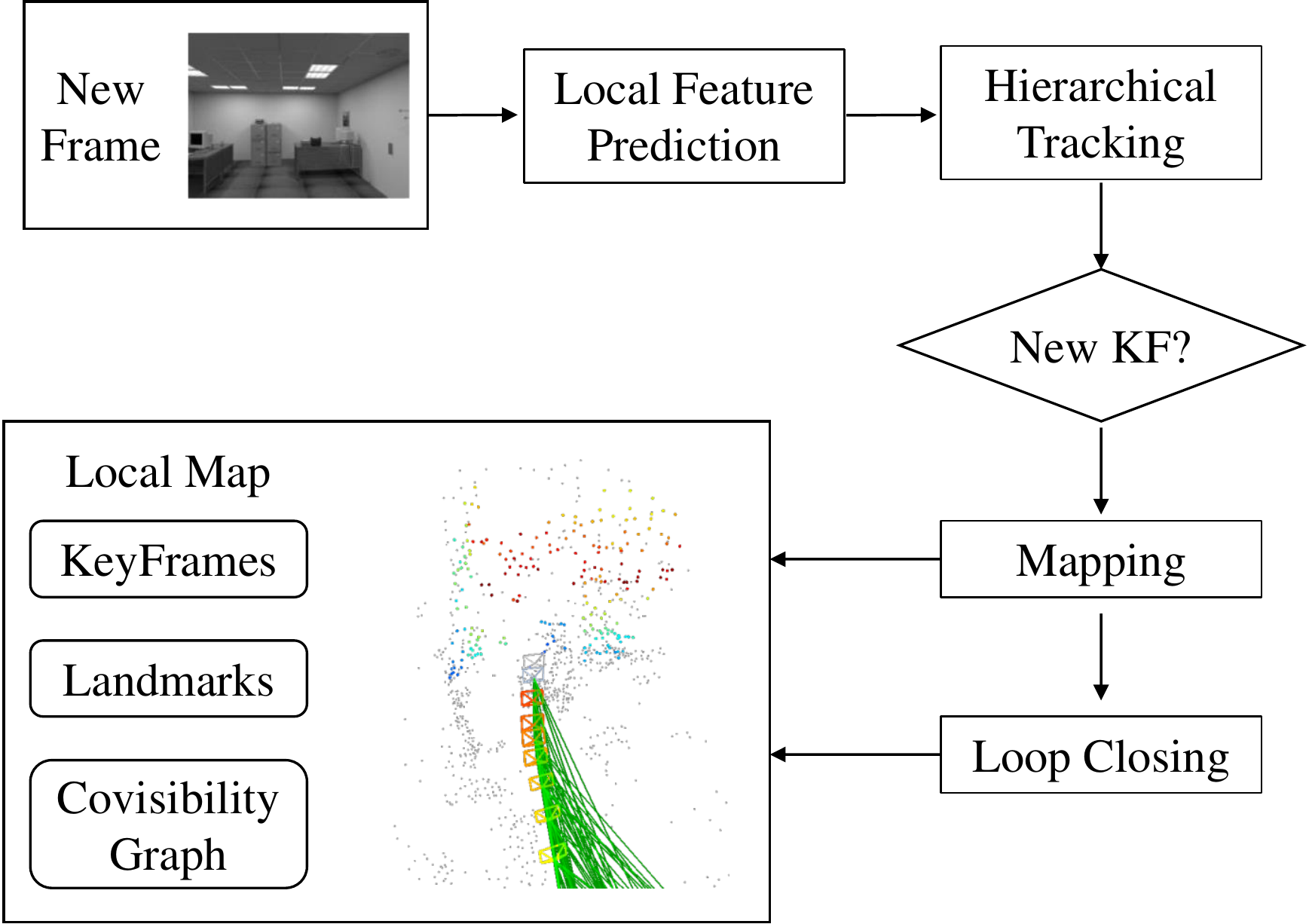}
    \caption{System overview. Our system leverages deep networks to predict both local and global features. We follow the systematic design of ORB-SLAM2 \cite{mur2017orb}, which can be divided into three modules: tracking, mapping and loop-closure. In the frontend, we propose a hierarchical scheme to recover the camera poses with network predictions. The mapping module generates new landmarks and maintains a covisibility graph. We further leverage learnt global embeddings to predict loop-closure and correct the global inconsistency.}
	\label{fig.system}
\end{figure}

Experimental results on public datasets, especially accurate pose estimations on challenging scenarios where state-of-the-art methods fail,
demonstrate the effectiveness of the proposed system.
\autoref{fig.system} provides an overview of our system.
Our contributions are summarized as follows:
\begin{itemize}
	\item
	      A monocular visual SLAM system leveraging learned local and global embeddings.
	\item 
	      A hybrid tracking scheme along with uncertainty modeling based on the network predictions.
	\item
	      A mapping module adapted from the traditional pipeline to fit the nature of the frontend.
	\item
	      Extensive evaluations on public datasets to demonstrate the accuracy and robustness of the proposed method.
\end{itemize}
Some parts of this work has been presented in \cite{hyhuang2020rdvo},   In this paper, we extend the conference paper \cite{hyhuang2020rdvo} with the following contributions:
\begin{itemize}
	\item We improve the camera tracking scheme and provide a probabilistic explanation.
	\item We extend the previous visual odometry system into a complete SLAM system with global features predicted by a neural network.
	\item We perform more experiments to verify the effectiveness of different modules and the overall system.
\end{itemize}

The rest of the article is organized as follows:
In \autoref{sec.review}, we review recent progress and challenges in VSLAM along with methods integrating deep learning.
In \autoref{sec.notation}, we demonstrate commonly used notations for simplicity.
Then, \autoref{sec.feature_extraction}, Sec. 5 and \autoref{sec.mapping} describe the three main modules in our system, namely feature extraction, camera tracking and association and mapping and loop closing, respectively.
In \autoref{sec.experiments}, we report the experimental results for the ablation and comparative studies.
Finally, \autoref{sec.conclusions} concludes this paper.

\section{Related Works}
\label{sec.review}

\subsection{Monocular Visual SLAM and its Challenges}
\label{sec.review_mvo}

The state-of-the-art approaches for monocular VO/VSLAM can be categorized into two dominant classes, namely \textit{indirect} and \textit{direct} methods. 
Indirect, or feature-based methods \cite{klein2007ptam, strasdat2011double, mur2015orb}, generally extract points of interest along with their descriptors as a sparse representation of the input image. 
With multiview correspondence association, camera motion and sparse structure are recovered via minimizing the reprojection error.
Among these works, ORB-SLAM \cite{mur2015orb} exploits the covisibility relationships to strengthen the map reuse and frame management, balancing the accuracy and the computational demand.

Direct methods \cite{newcombe2011dtam,engel2014lsd,engel2018dso}, on the contrary, directly minimize the photometric error on input images to track camera poses. 
In particular, DSO \cite{engel2018dso} optimizes camera poses, sparse scene structure, and camera model parameters jointly.
To combine the advantages of both methods, Froster \textit{et al}. propose SVO \cite{forster2014svo}, 
which tracks the camera poses via sparse image alignment and utilizes hierarchical bundle adjustment (BA) as the backend for optimizing the structure and camera motion.
Inspired by these works, we propose a hybrid scheme for the camera motion estimation, which tracks the pose initially on the predicted repeatability map and then refines it in an indirect manner.
In the context of challenging conditions, robust VO/VSLAM remains unsolved \cite{cadena2016past, park2017illumination, yang2018challenges}.
Potential solutions to these issues can be found in \cite{alismail2016direct, zhang2017active, kim2017robust}.
These works typically specialize in a particular problem (e.g., handling high dynamic range (HDR) environment), which would introduce certain overhead under common scenarios \cite{alismail2016direct}.
In contrast, our system is developed for a general VSLAM system, which does not require any preprocessing on the camera input for the image quality enhancement.
At the same time, in the experiments, it works well under most conditions.

\subsection{Deep Learning in Visual SLAM}

With the emergency of DL, recent researches demonstrate significant benefits by introducing relevant techniques into VSLAM systems.
Compared with traditional methods in local feature detection \cite{harris1988combined,shi1993good,rosten2006machine} and description \cite{lowe2004distinctive, bay2006surf, Rublee2011}, learning-based approaches \cite{detone2018superpoint,dusmanu2019d2,revaud2019r2d2, tang2019gcnv2,sarlin2019coarse} exhibit a competitive performance in both matching accuracy and efficiency.
Several works propose to use deep features in the camera pose estimation.
Tang \textit{et al}. \cite{tang2019ba} proposed BA-Net, where a feature-metric objective function is used for the direct image alignment, and the camera poses and predicted depth maps can be optimized in an end-to-end manner.
In \cite{gn-net-2020}, Stumberg \textit{et al}. proposed GN-Net, which introduced a Gauss-Newton loss for training the feature maps to be more invariant in the direct camera tracking.
Works attempting to combine learnt local features with VO/VSLAM have been brought to our view \cite{detone2018self,tang2019gcnv2}.
In \cite{detone2018self}, with labels from a VO backend, 
the original SuperPoint \cite{detone2018superpoint} is extended to predict the stability of local keypoints.
In \cite{tang2019gcnv2}, to enhance the frontend efficiency for onboard RGB-D VO, Tang \textit{et al}. proposed GCNv2.
They supervise the detector with ground-truth keypoint locations labeled by Shi-Tomasi score.
To further train the descriptor with triplet loss, positive and negative matches are retrieved via the projective geometry.
The above works generally focus on leveraging traditional methods (e.g., multiview geometry) to train a more practical and efficient frontend.
On the contrary, here we consider bridging the gap of exploiting learned features to alleviate the challenges in monocular VSLAM.
\section{Notations}
\label{sec.notation}

Throughout the paper, we denote the image collected at the $k$-th time as $I_k$ and the corresponding frame as $\mathcal{F}_k$.
The world frame $\mathcal{F}_w$ is set to be identical to the first camera frame $\mathcal{F}_0$.

For $I_k$, the rigid transform $\mathbf{T}_k \in \mathbf{SE}(3)$ maps a 3D landmark $\mathbf{x}_i \in \mathbb{R}^3$ in $\mathcal{F}_w$ to $\mathcal{F}_k$ using:
\begin{equation}
	^{c_k}\mathbf{x}_i = \mathbf{R}_k \mathbf{x}_i + \mathbf{t}_k,
\end{equation}
where $\mathbf{T}_k = \left[ \mathbf{R}_k | \mathbf{t}_k \right]$. $\mathbf{R}_k$ and $\mathbf{t}_k$ are the rotational and translational components of $\mathbf{T}_k$, respectively.
Accordingly, $^{c_k}\mathbf{x}_k$ denotes a 3D point in $\mathcal{F}_k$.

We denote a 2D pixel projected from a 3D landmark as $\tilde{\mathbf{u}}_{i, k}$. We use $\pi : \mathbb{R}^3 \rightarrow \mathbb{R}^2$ to denote the projection function: $\tilde{\mathbf{u}}_{i,k} = \pi(^{c_k}\mathbf{x}_{i})$, where $\tilde{\mathbf{u}}_{i, k}$ is the coordinate in the pixel coordinate. $\pi$ is defined as $\pi(^{c_k}\mathbf{x}_i)=\mathbf{K}{^{c_k}\mathbf{x}_i}$,
where $\mathbf{K}$ is the intrinsic matrix for pinhole model.

The update of a camera pose is parameterized as an incremental twist $\boldsymbol{\xi} \in \mathfrak{se}(3)$. We use a left-multiplicative formulation $\oplus : \mathfrak{se}(3)\times \mathbf{SE}(3)\rightarrow \mathbf{SE}(3)$ for the update of $\mathbf{T}_k$, which is denoted as:
\begin{equation}
	\boldsymbol{\xi}\oplus\mathbf{T}_{k}:=\exp ( \boldsymbol{\xi} ^{\land} )\cdot \mathbf{T}_k,
\end{equation}
where the mapping $\textbf{SE}(3) \rightarrow \mathbb{R}^6$ is defined by $\mathbf{T} = \exp (\boldsymbol{\xi}^\land)$,
where $(\cdot)^\land$ is the wedge operator.

\section{Learning-based Feature Extraction}
\label{sec.feature_extraction}

Here we adopt SuperPoint \cite{detone2018superpoint} as the feature extraction front-end.
Recall the pipeline of SuperPoint,
it first encodes the input image $I \in \mathbb{R}^{H\times W}$ with a single, shared encoder.
Then two different heads decode a repeatability volume $\mathcal{H}^{\prime} \in \mathbb{R}^{H_c \times W_c \times (C^2+1)}$ and a dense description $\mathcal{D}^{\prime} \in \mathbb{R}^{H_c \times W_c \times 256}$, respectively.
$C$ is the size of the grid cell, which is 8 in our system, and $H_c = H / C, W_c = W/C$.
$\mathcal{H}^\prime$ is then normalized by channel-wise softmax:
\begin{equation}
	\mathcal{H}(h_c, w_c, y)=\frac{\exp \left(\mathcal{H}^\prime(h_c, w_c, y)\right)}{\sum_{k=1}^{65} \exp \left(\mathcal{H}^\prime(h_c, w_c, k)\right)} .
\end{equation}
In this architecture, the keypoint detection problem is then formulated as a classification problem.
Specifically, for $\mathcal{H}(h_c, w_c, :) \in \mathbb{R}^{C^2 + 1}$,
the first $C^2$ channels represent the probability of a pixel in a $C \times C$ grid to be a keypoint, while the last channel stands for the probability that no interest point lies in the current patch.
For the details we refer the readers to \cite{detone2018superpoint}.

We define the repeatability map, in patch-wise ($\mathcal{R}_d$) and in pixel-wise ($\mathcal{R}$) as:
\begin{equation}
	\mathcal{R}_{d} = \mathcal{H}(., ., C^2 + 1), \enspace
	\mathcal{R} = -\log(s(\mathcal{H}(\cdot, \cdot, :-1))).
\end{equation}
where '.' denotes the omitted channels, while ':' denotes ``from the first channel to the last channel''. \textbf{s()} denotes the function that shuffles a feature vector into a 2D patch. In other words, it flattens the 3D volume to a 2D map.
Here we further define the patch location where the pixel $\mathbf{u}$ belongs to as $\mathbf{u}_p$.
Then a probabilistic explanation is given by:
\begin{equation}
	p(\mathbf{u} | I, \mathbf{w}) = \mathcal{R} (\mathbf{u}),
\end{equation}
\begin{equation}
	p(\mathbf{u}_p | I, \mathbf{w}) = 1 - \mathcal{R}_d (\mathbf{u}),
\end{equation}
where $\mathcal{R}_d \in \mathbb{R}^{H_c \times W_c}$, the last channel of $\mathcal{H}$, stands for nonexistence of interest point in current patch.
Accordingly, we reinterpret $\mathcal{R}_d$ as the patch-wise repeatability prediction, which is used for the initial camera tracking, described in \autoref{sec.tracking}.
$\mathcal{R} \in \mathbb{R}^{H\times W}$ is the repeatability map with full-resolution and $s: \mathbb{R}^{H_c \times W_c \times C^2}\rightarrow\mathbb{R}^{H \times W}$ maps the volume to a 2D heatmap.
In the above formulation, we negate the repeatability response, so that a pixel $\mathbf{u}$ (or patch $\mathbf{u}_\mathbf{p}$) is more leaning to be a keypoint (or to contain a keypoint) if it has a smaller response $\mathcal{R}(\mathbf{u})$ (or $\mathcal{R}_d(\mathbf{u}_{\mathbf{p}})$).

In a non-maximum suppression (NMS) scheme,
the locations of 2D features along with the final 2d grid $\mathcal{O}\in \mathbb{R}^{H_c\times W_c}$ are extracted from $\mathcal{R}$.
A single cell in $\mathcal{O}$ stores the index of the interest point in the current patch, zero if no salient point exists.
For a local feature $\mathbf{u}_i$, the corresponding descriptor $\mathbf{d}_i$ is sampled from $\mathcal{D} \in \mathbb{R}^{H\times W}$, which is interpolated from $\mathcal{D}^\prime$.
In addition, for the keyframes selected in our system, we predict the global embedding vector with NetVLAD \cite{arandjelovic2016netvlad}, which is used to measure the image similarity across different frames for detecting the loop closure.

\section{Camera Tracking and Association}

\subsection{Camera Pose Tracking}
\label{sec.tracking}
\begin{figure}[t!]
	\centering
	\includegraphics[width=0.5\textwidth]{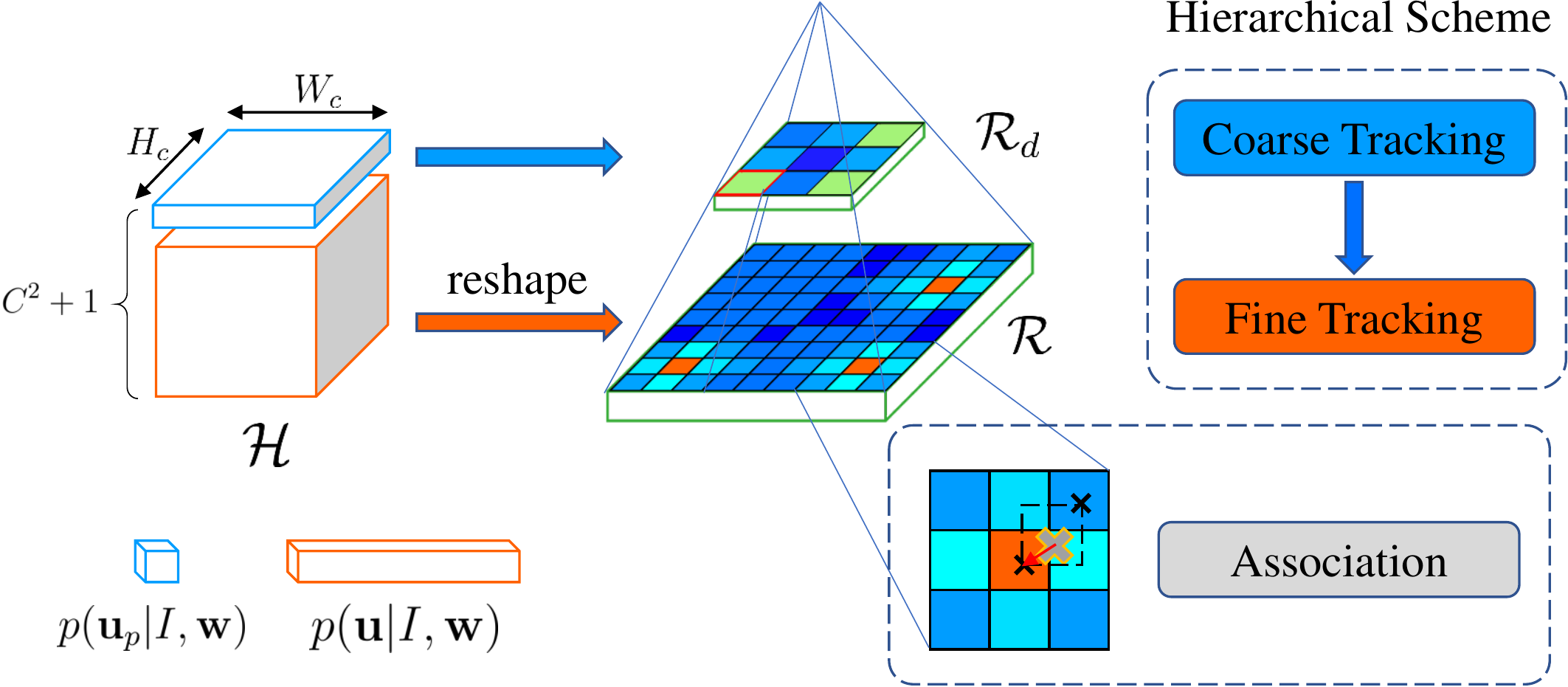}
	\caption{Tracking scheme of the proposed system. We reinterpret the local feature detection as a two-level feature probability heatmap. Our method then tracks the repeatability map in a coarse-to-fine manner. With the tracking result, we associate the features with landmarks locally.}
	\label{fig.tracking}
\end{figure}

Assuming an initial sparse structure built from the backend,
the frontend recovers the camera pose and associates 3D landmarks with 2D observations, which is illustrated in \autoref{fig.tracking}.

Initially, we track the current camera pose in a direct manner.
The direct method of the camera pose estimation advances in their correspondence-free formulation, where there is no need for heuristic design of matching strategies.
To take both the advantage of direct method and the network predictions,
we propose to hierarchically track the camera pose on the repeatability map.
Our intuition behind is that 3D landmarks triangulated from 2D observations are consistently repeatable in different views.
In other words, the reprojection location of a 3D landmark in $\mathcal{R}_{(\cdot)}$ should be the local extremum.

For the simplicity, here we denote the set of visible landmarks as $\mathcal{X} = \left\{\mathbf{x}_0, \mathbf{x}_1\dots \mathbf{x}_n \right\}$.
The current camera pose is denoted as $\mathbf{T}_k$ and is parameterized as $\boldsymbol{\xi}_k$.
Given a landmark position $\mathbf{x}_i$ and the current camera motion parameter $\boldsymbol{\xi}_k$, the likelihood of local observation can be derived as:
\begin{equation}
	p( \tilde{\mathbf{u}}_{ik} | \mathbf{w}, \boldsymbol{\xi}_k, \mathbf{x}_i) = \mathcal{R}(\tilde{\mathbf{u}}_{ik})= \mathcal{R}(\pi(\exp (\boldsymbol{\xi}_k^\land) \cdot \mathbf{x}_i)).
\end{equation}
To estimate $\boldsymbol{\xi}_k$, our method seeks to minimize the total likelihood, which is given by:
\begin{equation}
	\begin{split}
		\hat{\boldsymbol{\xi}}_k &= \arg \max_{\boldsymbol{\xi}_k} \prod_{\mathbf{x}_i \in \mathcal{X}} p(\cdot) = \arg \min_{\boldsymbol{\xi}_k} -\sum_{\mathbf{x}_i \in \mathcal{X}}\log(p(\cdot))\\
		& = \arg \min_{\boldsymbol{\xi}_k}
		-\sum_{\mathbf{x}_i \in \mathcal{X}}\log \left( \mathcal{R} (\pi(\exp (\boldsymbol{\xi}_k^\land) \cdot \mathbf{x}_i)) \right),
	\end{split}
	\label{eq.max_prob}
\end{equation}
where we abbreviate $p( \tilde{\mathbf{u}}_{ik} | \mathbf{w}, \boldsymbol{\xi}_k, \mathbf{x}_i)$ to $p(\cdot)$.
With a robust kernel function denoted as $\|\cdot\|_\gamma$, the total energy function is given by:
\begin{equation}
	E = \sum_{\mathbf{x}_i \in \mathcal{X}} \|-\log(\mathcal{R} (\pi(\exp (\boldsymbol{\xi}_k^\land) \cdot \mathbf{x}_i)))\|_\gamma.
\end{equation}
In the implementation, we choose Huber norm as the loss function.
Based on the above derivation, this problem is converted into a standard least-squares problem, which can be solved using Gauss-Newton method.

Direct method is considered very sensitive to the initial guess.
A common strategy to alleviate this issue is to leverage a pyramidal implementation.
In our case, generating pyramidal prediction results require multiple forward passes, prohibiting its real-time performance.
However, we disassemble the prediction into two levels, which represents the repeatability from coarse to fine, as described in \autoref{sec.feature_extraction} and shown in the \autoref{fig.tracking}.
Therefore, similar to the pyramidal implementation in the optical flow estimation or direct image alignment, we hierarchically track the current camera in a coarse-to-fine manner.
For estimating the camera pose in the coarsest level, oppositely we minimize the probability that a patch does not contain keypoint, causing a little subtlety in the formulation. This gives:
\begin{equation}
	\begin{split}
		\hat{\boldsymbol{\xi}}_k &= \arg \min_{\boldsymbol{\xi}_k} p(\mathbf{u}_p | \boldsymbol{\xi}_k, \mathbf{x}_i, \mathbf{w}), \\
		& = \arg \min_{\boldsymbol{\xi}_k} \mathcal{R}_d (\pi_d(\exp (\boldsymbol{\xi}_k^\land) \cdot \mathbf{x}_i)) .
	\end{split}
	\label{eq.max_prob}
\end{equation}

\begin{algorithm}[t]
	\providecommand{\DontPrintSemicolon}{\dontprintsemicolon}
	\KwData{$I_{k}, \tilde{\boldsymbol{\xi}}_k, \mathcal{O}_k, \mathcal{R}, \mathcal{R}_d, \mathcal{M}$.}

	$\tilde{\boldsymbol{\xi}}_k$ = applyConstantVelocity($\tilde{\boldsymbol{\xi}}_k$)

	$\mathcal{X}$ = collectVisibleLandmarks($\mathcal{M}$)

	\CommentSty{//$\mathcal{M}$ denotes the local map.}

	$\hat{\boldsymbol{\xi}}_k$ = coarseTracking($\tilde{\boldsymbol{\xi}}_k$, $\mathcal{X}$)

	\ForEach(){$\tilde{\mathbf{u}}_{i,k}$}{

		$\mathcal{V}_{i,k}$ = checkAdjacentPatches($\tilde{\mathbf{u}}_{i,k}$, $\mathcal{O}_k$)

		\eIf(){$|\mathcal{V}_{i,k} | \equiv 1$}{
			associate($\mathbf{x}_{i}$, $\mathcal{V}_{i,k}[0]$)
		}()
		{
			$j$ = findBestAssociation($\mathbf{x}_{i}$, $\mathcal{V}_{i,k}$)

			associate($\mathbf{x}_{i}$, $\mathcal{V}_{i,k}[j]$)
		}

	}

	$\hat{\boldsymbol{\xi}}_k$ = poseRefinement($\hat{\boldsymbol{\xi}}_k$, $\mathcal{X}$)

	outlierFiltering($\hat{\boldsymbol{\xi}}_k$, $\mathcal{X}$)

	\caption{Tracking and Feature Association}
	\label{alg.tracking}
\end{algorithm}

\begin{figure}[t!]
	\centering
	\subfloat{
		\includegraphics[width=0.2\textwidth]{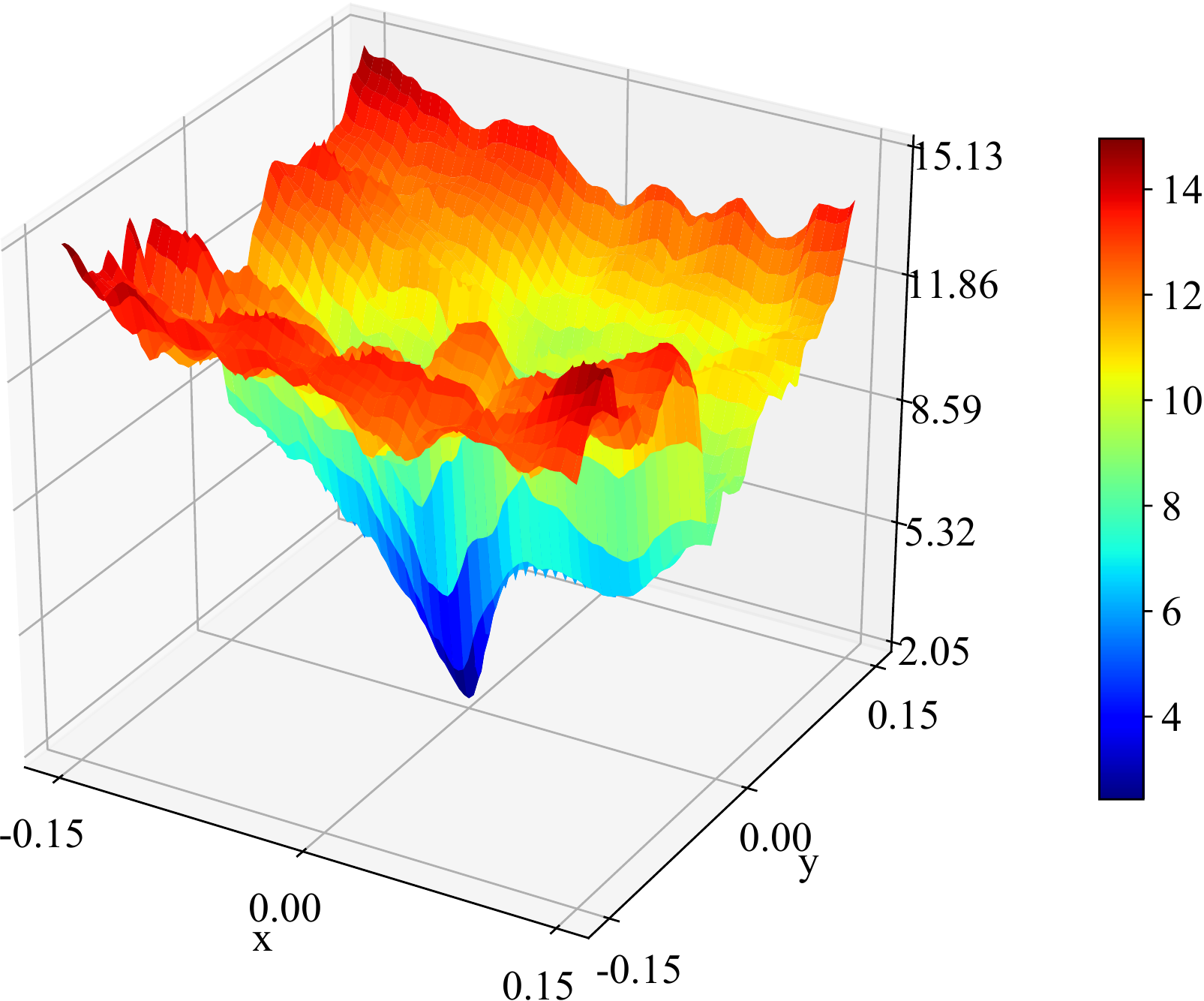}
	}\quad
	\subfloat{
		\includegraphics[width=0.2\textwidth]{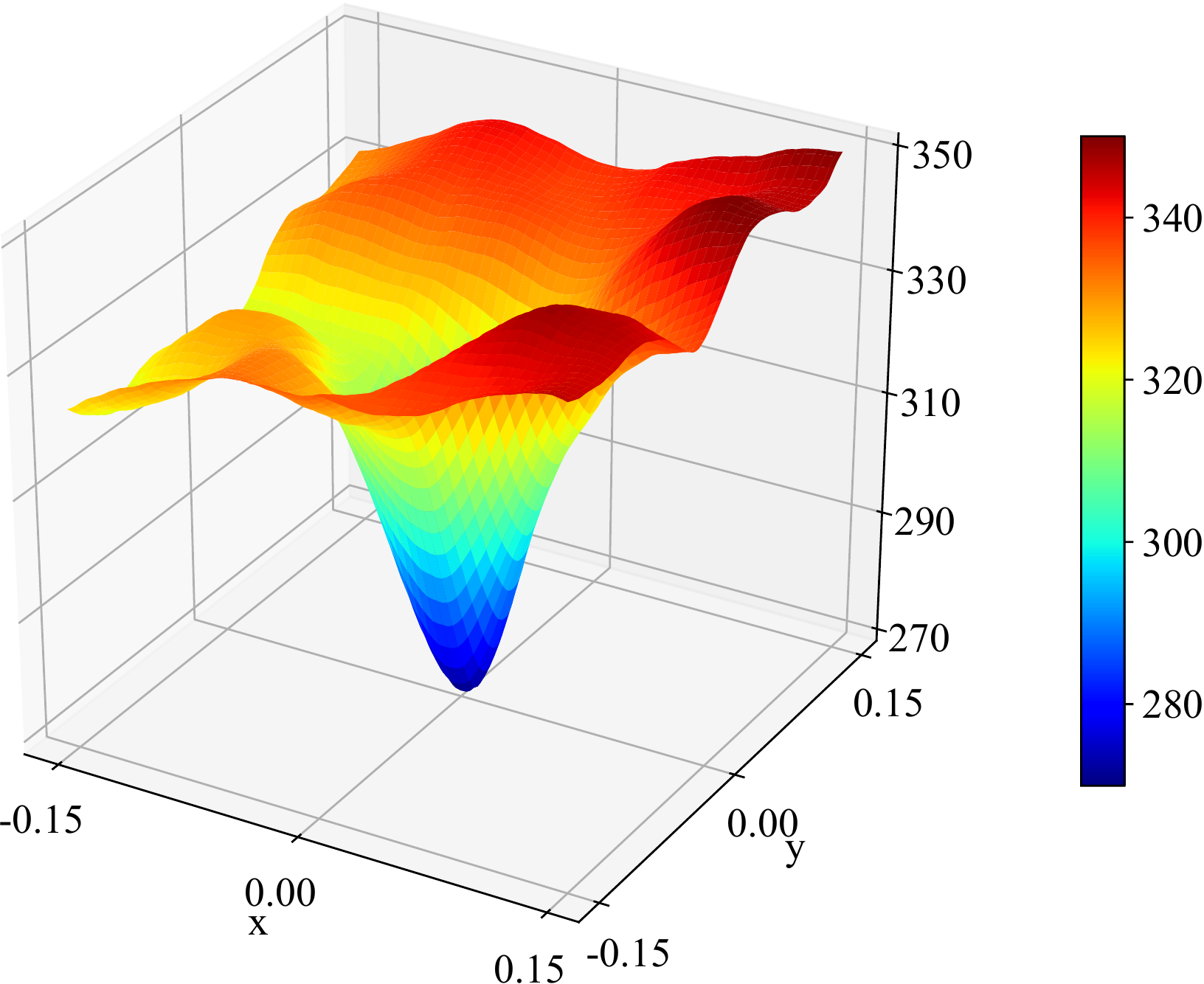}
	}
	\caption{Cost surface for camera pose tracking of \textit{photometry}-based residual (left) and \textit{repeatability}-based residual (right).
		For each plot, the x-y plane stands for different translational offsets to the ground-truth pose and the value on z-axis is the total cost with the corresponding transform.}
	\label{fig.track_cost}
	\vspace{-0.3cm}
\end{figure}

For each residual term, the jacobian can be derived as:
\begin{equation}
	\begin{split}
		\mathbf{J}_i & = \mathbf{J}_\text{repeat} \cdot \mathbf{J}_{\text{proj}} \cdot \mathbf{J}_{\text{pose}}\\
		& = \frac{\partial \mathcal{R}_{(\cdot)}( \tilde{\mathbf{u}}_{i, k} )}{\partial \tilde{\mathbf{u}}_{i, k}}
		\cdot \frac{\partial \tilde{\mathbf{u}}_{i, k}}{\partial \mathbf{x}_i^c}
		\cdot\frac{\partial \left(\left(\boldsymbol{\xi}_k \oplus \mathbf{T}_k \right)\mathbf{x}_i\right)}{\partial\boldsymbol{\xi}_k}, \\
	\end{split}
\end{equation}
where similar to direct tracking methods, $\mathbf{J}_\text{repeat}$ is the gradient of the corresponding repeatability map, which is evaluated at the projected pixel $\tilde{\mathbf{u}}_{i, k}$ and is calculated by bilinear interpolation at each iteration.
$\mathbf{J}_\text{proj}$ and $\mathbf{J}_\text{pose}$ are the Jacobian of projection function w.r.t transformed point and the left-compositional derivative of the the transformed point w.r.t the twist update $\boldsymbol{\xi}_k$, respectively.

In practice, we find that tracking on the coarse level already produces a relatively good pose estimation result. When this estimation is fed to the fine level, our tracking module is able to fine-tune it for better accuracy. Analogous to the traditional optical flow pipeline, the pyramidal implementation is a significant design to alleviate the convergence basin issue and make the tracking module work better with predicted repeatability.
\autoref{fig.track_cost} compares the cost surfaces of our method and the classical direct method.
The cost surface of photometry-based tracking suffers from non-convexity responsible for the sensitivity to initial guess, the narrowness of convergence basin, and potential numerical issues in the optimization. 
In contrast, tracking on repeatability exhibits a smoother cost surface, which in practice yields better convergence property.

\subsection{Feature Association}

With the initial tracking, Here we propose a local feature association scheme, which minimizes the computational overhead by avoiding the descriptor matching as much as possible.
Our observation is that, with the initial tracking, the reprojection location is considered to have an accuracy at sub-pixel level.
That is to say, if the optimization converges, the reprojection location $\tilde{\mathbf{u}}_{i, k}$ of an inlier $\mathbf{x}_i$ is supposed to have sub-patch accuracy.
In addition, as we described before, we perform NMS or griding to store one feature in a $8\times 8$ patch.
Under these observations, the 2D association of $\mathbf{x}_i$ should belonging to the four adjacent patches or grid cells of $\tilde{\mathbf{u}}_{ik}$.
Therefore, instead of exhaustively matching descriptors or adopting some heuristics for tracking, we actively attempt to associate features in four adjacent patches.
Based on the actual number of adjacent patches which contain a keypoint, we associate landmarks and local features as follows:
\begin{itemize}
	\item If there is only one keypoint candidate, we directly associate it with the landmark.
	\item If there are multiple keypoint candidates, we then associate it according to descriptor distance, which is equivalent to finding the best local association.
\end{itemize}
It is noticeable that this association step is achieved with minimal computational overhead, which might be caused by matching high-dimensional descriptors.

\subsection{Feature Parameterization and Pose Refinement}
For the indirect methods, the \textit{reprojection error} is generally used for pose estimation or Bundle Adjustment. A generic formulation is given by:
\begin{equation}
	\mathbf{u}_{i, k} = \mathbf{h}_{i, k}(\mathbf{x}_i, \boldsymbol{\xi}_k) + \mathbf{n}_{i, k},
\end{equation}
where $\mathbf{h}_{i, k}$ is the measurement function between the $i$-th landmark position $\mathbf{x}_i$ and $k$-th keyframe pose $\boldsymbol{\xi}_k$; and $\mathbf{n}_{i, k}$ is zero-mean white Gaussian noise that perturbs the measurement, i.e., $\mathbf{n}_{i, k} \sim \mathcal{N}(\mathbf{0}, \boldsymbol{\Sigma}_{i, k})$.
In other words, this formulation models local observations as 2D Gaussian components, i.e., $\tilde{\mathbf{u}}_{i, k} \sim \mathcal{N}(\mathbf{\bar{u}}_{i,k}, \boldsymbol{\bar{\Sigma}}_{i, k})$.
As long as the majority of methods follow this generic formulation, the problem is how to define the parameters for each local feature, i.e., $\mathbf{\bar{u}}_{i, k}$ and $\boldsymbol{\bar{\Sigma}}_{i, k}$.
Previous works resolve this issue in two approaches.
In the early stage, methods model local features as 2D Gaussian distributions with anisotropic covariance.
Some interesting discussions can be found in \cite{kanazawa2003we, brooks2001value}, where the authors generally conclude modelling such anisotropic covariance can improve the visual state estimation.
On the contrary, modern SLAM systems \cite{mur2015orb,mur2017orb,forster2014svo} generally assign an isotropic covariance for each feature.
This gives $\boldsymbol{\bar{\Sigma}}_{i, k} = \sigma^2 \mathbf{I}_{2\times 2}$, with $\sigma$ proportional to the pyramidal level of the corresponding feature.
However, as described above, generating a pyramidal prediction is prohibitive, making this parameterization method not practical in our implementation.

Following previous works, we seek to weigh the contribution of different local features to the overall estimation.
Compared to traditional feature extraction techniques, CNNs have a larger receptive field and are capable of generating more consistent local features.
Therefore, we model their 2D distribution from keypoint prediction heatmap with a probabilistic explanation.
Considering the prediction as a mixture of multivariate Gaussian distributions, to approximate the single variate parameters, we have:
\begin{equation}
	\bar{\mathbf{u}} = \mathbb{E}[\mathbf{u}] = \sum_{\mathbf{u} \in \mathcal{P}}\mathbf{u} \cdot p(\mathbf{\mathbf{u}} | \mathbf{w})
\end{equation}
\begin{equation}
	\begin{split}
		\bar{\mathbf{\Sigma}} &= \mathbb{E}[(\mathbf{u} - \bar{\mathbf{u}})(\mathbf{u} - \bar{\mathbf{u}})] \\
		& = \sum_{\mathbf{u} \in \mathcal{P}} (p_i^2 \mathbf{\Sigma}_{\mathbf{u}} + p_i \mathbf{u}_i \mathbf{u}_i^T) - \mathbf{u} \mathbf{u}^T\\
	\end{split}
\end{equation}
For the simplicity of implementation, here we represent $\mathcal{P}$ as a $3 \times 3$ patch.
This parameterization scheme is further verified by \autoref{sec.eval_param}.

After associating landmarks with the local observations, the current camera pose $\mathbf{T}_k$ is refined by minimizing the reprojection error.
For $\mathbf{x}_i$, the error function is defined as:
\begin{equation}
	\mathbf{e}_{i, k}^{\text{repro}}=\pi\left(\mathbf{R}_{k} \mathbf{x}_{i}+\mathbf{t}_{k}\right)- \tilde{\mathbf{u}}_{i, k}.
\end{equation}
Similar to the initial tracking, the refined pose is solved by iterative least-squares. The overall energy function to minimize is defined as:
\begin{equation}
	\label{eq:motion_ba}
	E^\text{repro} = \sum_{i \in \mathcal{P}_k}\|\left(\mathbf{e}_{i, k}^{\text{repro}}\right)^T \boldsymbol{\Sigma}_{i,k}^{-1} \mathbf{e}_{i, k}^{\text{repro}}\|_\gamma,
\end{equation}
where $\boldsymbol{\Sigma}_{i,k}$ is the covariance of 2D feature location $\tilde{\mathbf{u}}_{i, k}$ for weighting different features' contribution to the optimization.
The total pipeline of our tracking and association module is also demonstrated in \autoref{alg.tracking}.

\section{Mapping and Loop Closing}
\label{sec.mapping}

\subsection{Mapping}

\subsubsection{Landmark Creation}

When a new keyframe is constructed, the mapping module first creates new landmarks with previous observations.
For correspondence search, the top-$n$ keyframes sharing the most covisibility score with the current keyframe are chosen to be the candidate frames.
ORB-SLAM2 \cite{mur2017orb} exploits bag-of-words (BoW) \cite{GalvezTRO12} to accelerate feature matching between keyframes for the landmark generation.
Inevitably, this strategy introduces quantization effect that fewer correspondences would be found compared to brute-force search.
For original ORB-SLAM2, we observe that such effect is affordable as it extracts quantities of features (e.g., 1000 for a image with VGA resolution).
However, in our case, two factors amplify this quantization effect:
first, we only extract the most representative features, limiting the number of total local features;
second, float descriptors lie in a much larger space than binary descriptors, making quantization effect more significant as more clusters are required for a representative quantization.
This degrades both the robustness and the accuracy in practice.
Therefore, two methods are adopted for feature association between keyframes.
The first one is the approximate nearest neighbor (ANN) search.
At the creation of each keyframe, a database of untracked local features is established.
In the association step, we query each feature from the databases of candidate keyframes.
Moreover, after the association step, the database is reindexed to guarantee the search efficiency with future keyframes.
The problem of ANN-based association is that for the repetitive pattern (e.g. checkboard), ambiguity exists and increases the number of outliers in the triangulation step.
\begin{figure*}[t!]
	\centering
	\includegraphics[width=0.47\textwidth]{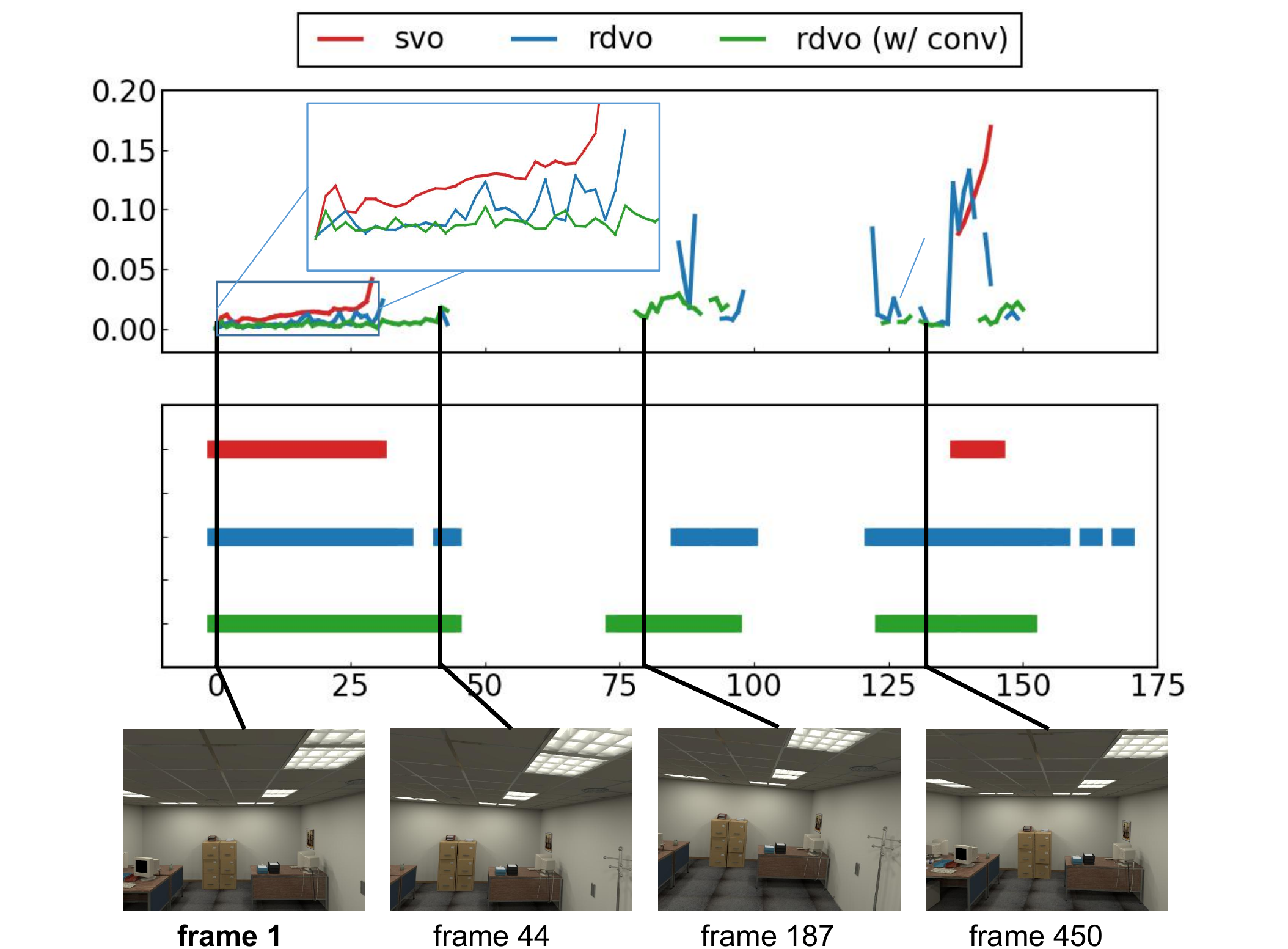}
	\includegraphics[width=0.47\textwidth]{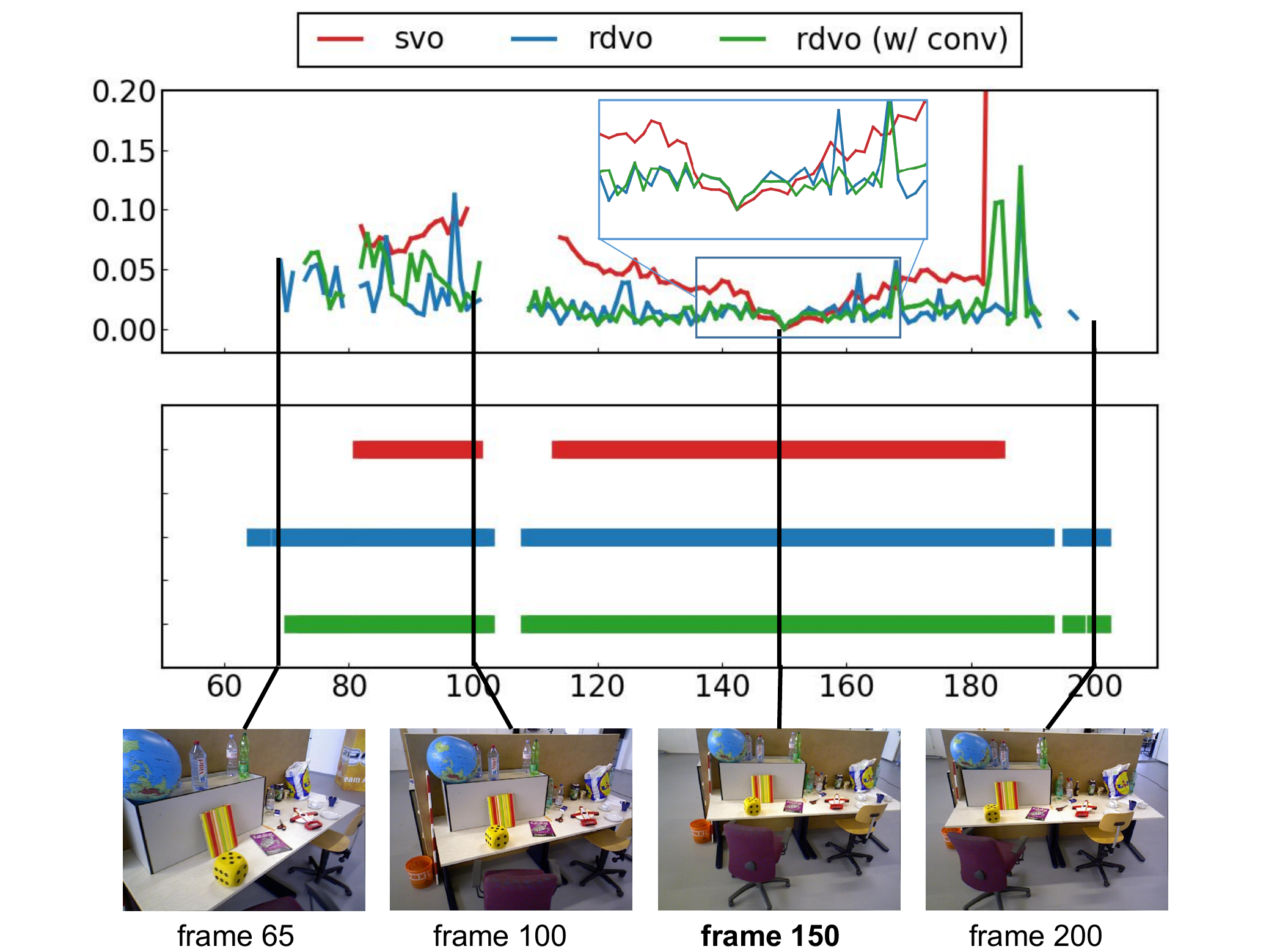}
	\caption{ Convergence radius and accuracy of different direct image alignment methods (indicated by color).
		All frames of the respective sequence are tracked on frame 1 (left) and frame 150 (right), using the identity as initialization.
		The bottom plots show for which frames tracking succeeds; the top plots show the final translational error.
		We observe that apart from the traditional photometry-based direct tracking (svo), the proposed repeatability-based tracking advances in larger convergence basin and better tracking precision.
	}
	\label{fig.eval_direct}
	\vspace{-0.1cm}
\end{figure*}

To resolve this problem, we further search the epipolar line of the 2d grid $\mathcal{O}_j$ of the target keyframe $\mathcal{F}_j$ to associate the features.
For an untracked local feature $\mathbf{u}_{i, k}$ in the current keyframe $\mathcal{F}_k$, the epipolar line $^i\mathbf{l}_{k, j} = \left[l_0, l_1, l_2\right]^T$ on the image plane of $\mathcal{F}_j$ is derived as:
\begin{equation*}
	(^i\mathbf{l}_{k,j})^T=\mathbf{u}_{i,k}^T\mathbf{K}^{-T} \mathbf{t}_{k,j}^{\land}\mathbf{R}_{k,j}
	\mathbf{K}^{-1},
\end{equation*}
\begin{equation*}
	\text{with} \quad\mathbf{R}_{k, j} = \mathbf{R}_k \mathbf{R}_j^{-1} \quad \mathbf{t}_{k, j} = - \mathbf{R}_k \mathbf{R}_j^T \mathbf{t}_j + \mathbf{t}_k,
\end{equation*}
where $\mathbf{K}$ is the projection matrix.
We search $\mathcal{O}_j$ along the entire epipolar line $^i\mathbf{l}_{k,j}$.
If current grid has an unmatched feature, we further check the epipolar distance $d_{i,k,l,j}$ as the geometry constraint:
\begin{equation*}
	d_{i,k,l,j} = \frac{1}{\text{det}(\boldsymbol{\Sigma}_{i,k})}\frac{\mathbf{u}_{l,j}^{T} \cdot {^i}\mathbf{l}_{k,j}}{\sqrt{l_0^2 + l_1^2}},
\end{equation*}
which is weighed by $1/\text{det}(\boldsymbol{\Sigma}_{i,k})$ for the consideration of uncertainty.
The inlier with the best descriptor distance is considered as a successful match for the candidate feature.
Finally, the 3D positions are recovered via mid-point triangulation.
For rejecting outliers in the triangulation, we follow \cite{mur2015orb} to further verify the sign of depth and the reprojection error in both keyframes.

\subsubsection{Backend Optimization}
Similar to ORB-SLAM2, batched BA is utilized for backend optimization and map management.
The variables in the local map for optimization consisting of updates to keyframe poses $\mathcal{T} = [\boldsymbol{\xi}_{k_0}, \boldsymbol{\xi}_{k_1}, ... , \boldsymbol{\xi}_{k_n}]$
and positions of landmarks $\mathcal{M} = [\mathbf{x}_{i_0}, \mathbf{x}_{i_1}, ... , \mathbf{x}_{i_m}]$.
We denote the full state vector as $\mathcal{X} = [\mathcal{T}, \mathcal{M}]$, which is solved by:
\begin{equation}
	{\mathcal{X}}
	= \arg \min_{\mathcal{X}}
	\sum_{\mathbf{T}_k}
	\sum_{\mathbf{x}_i}
	\text{obs}(i,k)
	\|(\mathbf{e}_{i, k}^{\text{repro}})^T \boldsymbol{\Sigma}_{i,k}^{-1} \mathbf{e}_{i, k}^{\text{repro}}\|_\gamma,
\end{equation}
where $\text{obs}(i,k) = 1$ if $\mathbf{x}_i$ is observed by $\mathcal{F}_k$, $\text{obs}(i,k)=0$ otherwise.
Note that, as in ORB-SLAM2 \cite{mur2017orb}, the poses of keyframes that share no observations of visible landmarks with the current keyframe are fixed to maintain a consistent scale estimation.
The map management operations are inherited from \cite{mur2017orb}. Briefly, it culls redundant keyframes, removes outliers in the local BA and fuses the landmarks. 

\subsection{Loop Closing}
We extend the built-in loop-closure module to work with learnt global embeddings.
When a keyframe is created, we generate the global embedding from NetVLAD \cite{arandjelovic2016netvlad}.
The advantage of using traditional BoW is that we can leverage the inverted file formulation for efficient keyframe retrieval.
Yet in our case, the network directly outputs a global embedding vector instead of preserving the visual words information.
Therefore, in our implementation, we directly compare the $l2$-distance between the global embeddings between the current keyframe and the past keyframes that does not share any common observations with the current keyframe.
Then loop-closure candidates are considered to have embeddings close to that of the current keyframe.
With the loop closure candidates, we further adopt the built-in modules of ORB-SLAM2 \cite{mur2017orb} to perform candidate verification and loop closure correction.
Briefly, if the loop closure is detected successfully, it first corrects the $\textbf{Sim}(3)$ pose graph generated by the minimal spanning tree from the global covisibility graph.
Then, global BA is performed with a similar pipeline described in \autoref{sec.mapping}.

\section{Experimental Results}
\label{sec.experiments}

In this section, we first perform ablation studies on different modules in our system, including the proposed direct tracking frontend and the feature parameterization.
Then, we report the general state estimation accuracy on several public datasets and compare our system against other popular methods.
Finally, we provide several qualitative results along with runtime analysis to further demonstrate the effectiveness of the proposed method.

\subsection{Evaluation on the Camera Tracking}
In this section, we validate the proposed direct tracking scheme.
For the baselines to compare with, we first select the front-end of SVO \cite{forster2014svo}.
The tracking front-end of SVO detects FAST corners in a grid,
and tracks recovered 3D landmarks in a sparse image registration manner,
which shares a similar scheme to our system and is considered to be a good competitor.
In addition, inspired by the recently proposed GN-Net in \cite{gn-net-2020},
we propose another baseline as competitor,
which tracks the current frame with deep features as input in a direct manner. The objective function is given by:
\begin{equation}
	E_\text{feat} = \sum\|\mathbf{F}_{k^\prime}( \mathbf{u}_{i, k^\prime}) - \mathbf{F}_k(\pi(\mathbf{R}_k \mathbf{x}_i + \mathbf{t}_k))\|,
\end{equation}
where $\mathbf{F}_{k}$ is the feature map of the $k$-th frame.
This formulation is based on the insight that CNNs can produce invariant feature for the correspondence across different frames.
Similarly, we add this \textit{feature-metric constraint} into our repeatability formulation.
In the implementation, we leverage the feature volume predicted by SuperPoint and reduce its dimension using PCA to achieve an applicable real-time implementation on CPU.
We denote this baseline as \textbf{rdvo (w/ conv)} and our method as \textbf{rdvo}.

Here we mainly evaluate the convergence radius and pose estimation accuracy of the tracking module.
For the evaluation, given a test sequence, we first generate a sparse pointcloud with a pair of RGB image and depth map.
Then, we evaluate different methods by registering the sparse visual structure to each frame in the sequence.
The experiments are conducted on the ICL-NUIM \cite{handa2014icl} and TUM Dataset \cite{sturm12iros}.
The results are shown in \autoref{fig.eval_direct}.
We observe that compared to SVO, the proposed method generally enlarges the convergence radius for direct tracking.
In other words, starting from the same initialization point, more frames with larger translation and rotation variances are successfully registered by our method.
This confirms the effectiveness of the proposed direct tracking scheme with predicted repeatability.

It is also interesting that, introducing high-level descriptions from the convolutional feature, rdvo (w/ conv) provides a close localization accuracy and robustness against rdvo.
While both of them outperform the tracking scheme in SVO, rdvo (w/ conv) enlarges the convergence radius a little on the ICL-NUIM dataset, which indicates that using the \textit{feature-metric} constraints can increase the robustness of direct tracking to some extend.
Despite that the convergence radius is increased as expected, we notice that, rdvo (w/ conv) does not consistently outperform rdvo and svo.
We think the reason could be that the image alignment generally requires a subpixel accuracy for an accurate alignment result.
Unlike repeatability predictions that implicitly provide pixel-level keypoint location,
convolutional feature slices of correspondences in different frames and have some subtleties with viewpoint variances.
These subtleties yield some degradation to the registration results.
Besides, although we interpolate between feature slices, the convolutional feature map is considered to provide patch-level information, which is downsampled 8 times in our case.
As the residual dimension is very large (256 in our case), implementing it on CPU leads to significant computational overhead, especially for interpolating the feature map.

There exist some image alignment methods based on raw gradient input \cite{ling2018edge,zhou2018canny,nurutdinova2015towards,kuse2016robust}.
However, these methods differs from baselines in the experiments in that they densely select pixels with high gradient and generally transform the gradient map into another space, e.g., Distance Transform (DT), for smoother gradient information \cite{zhou2018canny, kuse2016robust}.
Some also assume structural constraints between landmarks for better structural representation \cite{engel2014lsd}.
As a consequence, we only focus on \textit{sparse} image alignment methods here.

\begin{figure}[t!]
	\centering
	\includegraphics[width=0.48\textwidth]{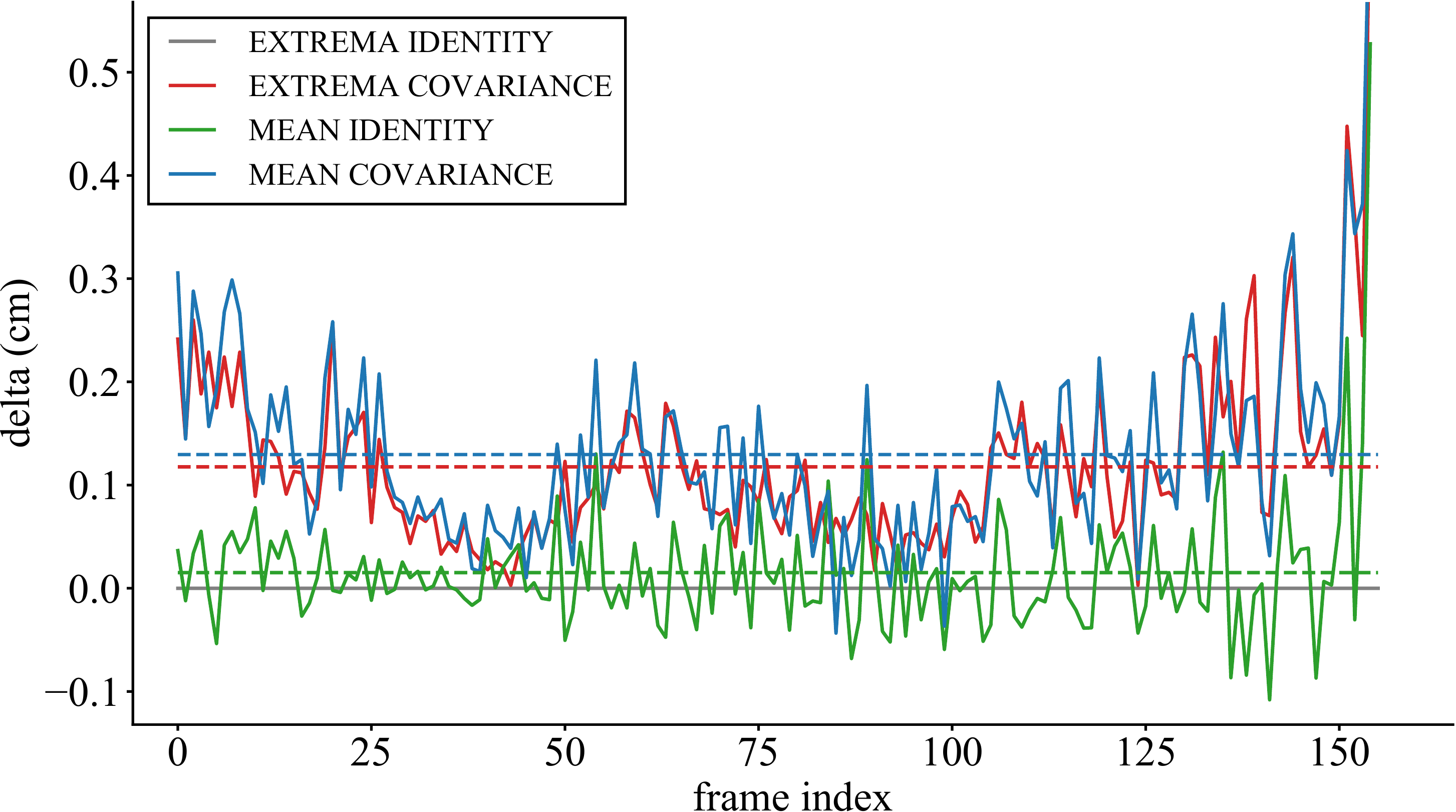}
	\caption{
		Relative improvement on different methods compared to the baseline method. The dashed line represents mean error of the corresponding method.
	}
	\label{fig.eval_param}
	\vspace{-0.1cm}
\end{figure}

\begin{table}[t!]
	\caption{Translational RMSE (cm) and tracking success rate (\%) on the New Tsukuba dataset. The tracking success is defined as (\#tracked frame/\#frame in the sequence).}
	\label{tab:tsukuba_acc}
	\centering
	\begin{tabular}{cccc}
		\toprule

		\multirow{2}{*}{\textbf{Sequence Name}} & \multirow{2}{*}{\textbf{Ours}} & \multirow{2}{*}{\textbf{DSO}} & \textbf{ORB-SLAM}     \\
		                                        &                                &                               & \textbf{w/o loop}     \\
		\midrule

		\multirow{2}{*}{fluorescent}            & \textbf{10.9 $\pm$  5.7}       & 59.0  $\pm$ 39.0              & (18.2 $\pm$ 15.3)     \\
		                                        & 87.6 $\pm$  0.3                & 98.2 $\pm$  0.0               & 88.4 $\pm$ 11.6       \\ \cline{2-4}\Tstrut{}
		\multirow{2}{*}{daylight}               & \textbf{9.4  $\pm$  4.9}       & 50.0  $\pm$ 29.7              & 14.7 $\pm$  9.7       \\
		                                        & 96.2 $\pm$  0.2                & 98.1 $\pm$  0.0               & 82.4 $\pm$ 16.9       \\ \cline{2-4}\Tstrut{}
		\multirow{2}{*}{lamps}                  & \textbf{9.3  $\pm$  4.6}       & $\boldsymbol{\times}$         & $\boldsymbol{\times}$ \\
		                                        & 94.5 $\pm$  7.1                & 44.5 $\pm$  37.4              & 0.0 $\pm$ 0.0         \\ \cline{2-4}\Tstrut{}
		\multirow{2}{*}{flashlight}             & \textbf{18.3 $\pm$  10.3}      & $\boldsymbol{\times}$         & $\boldsymbol{\times}$ \\
		                                        & 94.4 $\pm$  0.1                & 58.4 $\pm$  10.9              & 0.0 $\pm$ 0.0         \\

		\bottomrule
	\end{tabular}
\end{table}

\begin{figure*}[t!]
	\centering
	\subfloat[]{\includegraphics[trim={20px 0 120px 0},clip,width=0.25\textwidth]{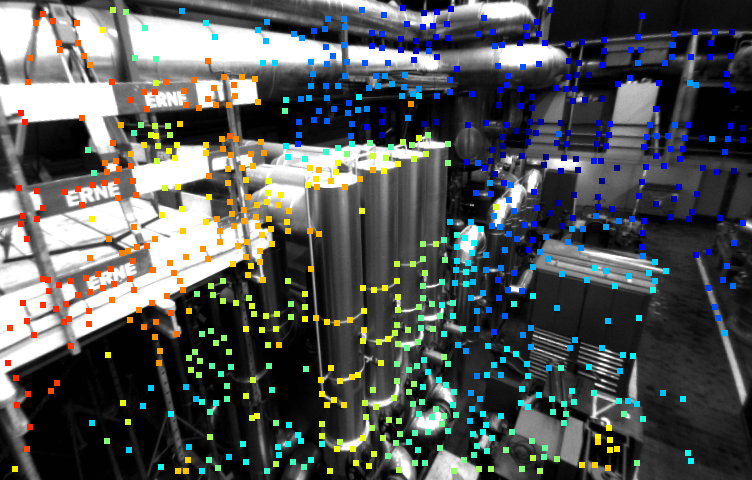}} \hspace*{0.001em}
	\subfloat[]{\includegraphics[trim={70px 0 70px 0},clip,width=0.25\textwidth]{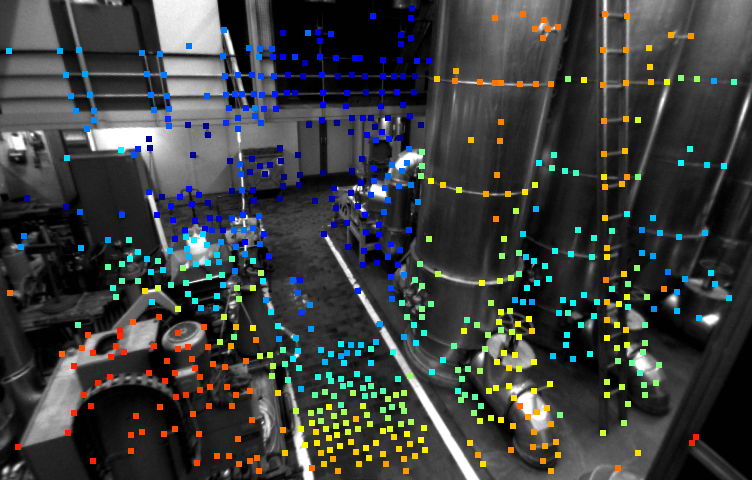}} \hspace*{0.001em}
	\subfloat[]{\includegraphics[height=0.28\textwidth]{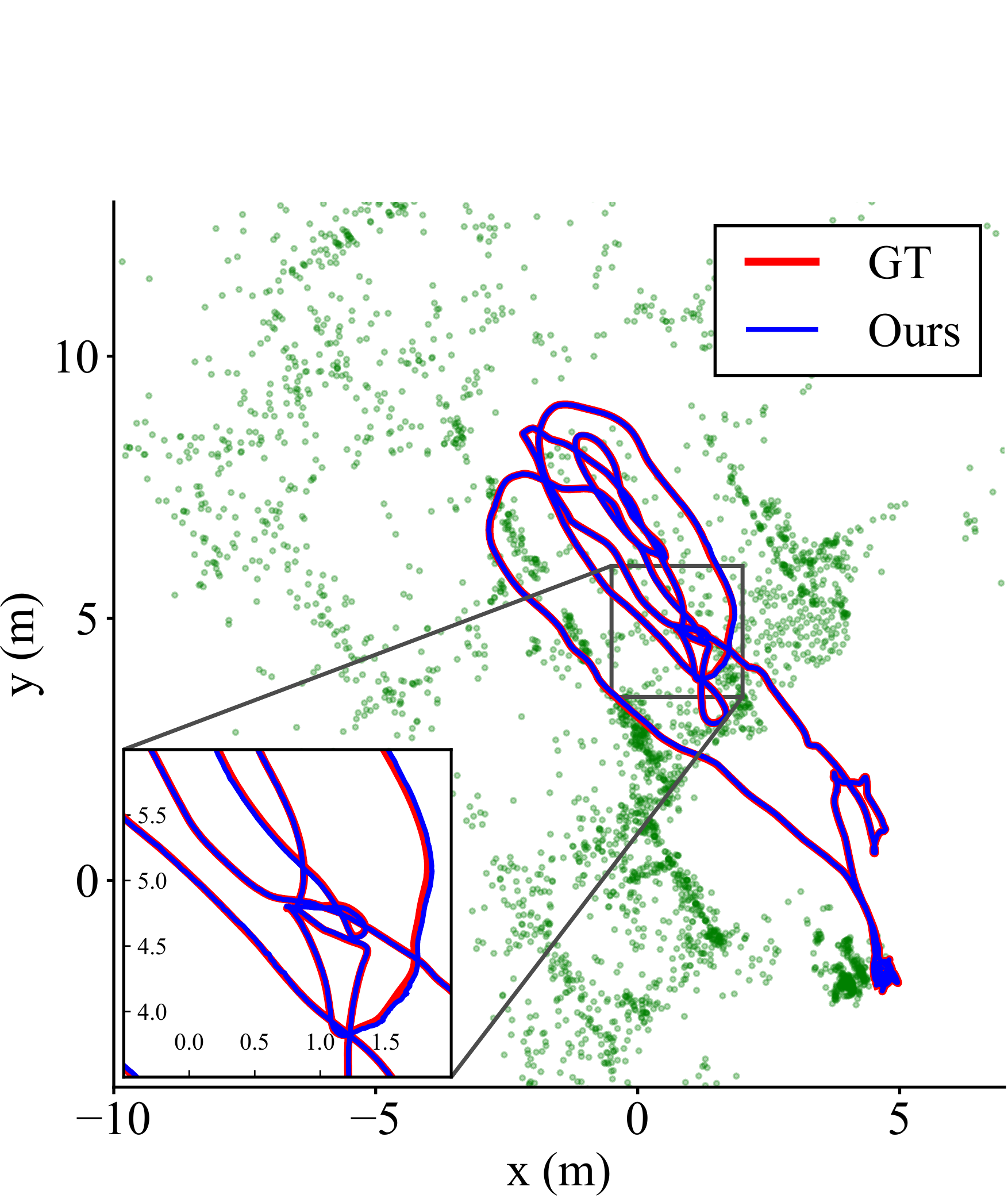}} \hspace*{0.001em}
	\subfloat[]{\includegraphics[height=0.28\textwidth]{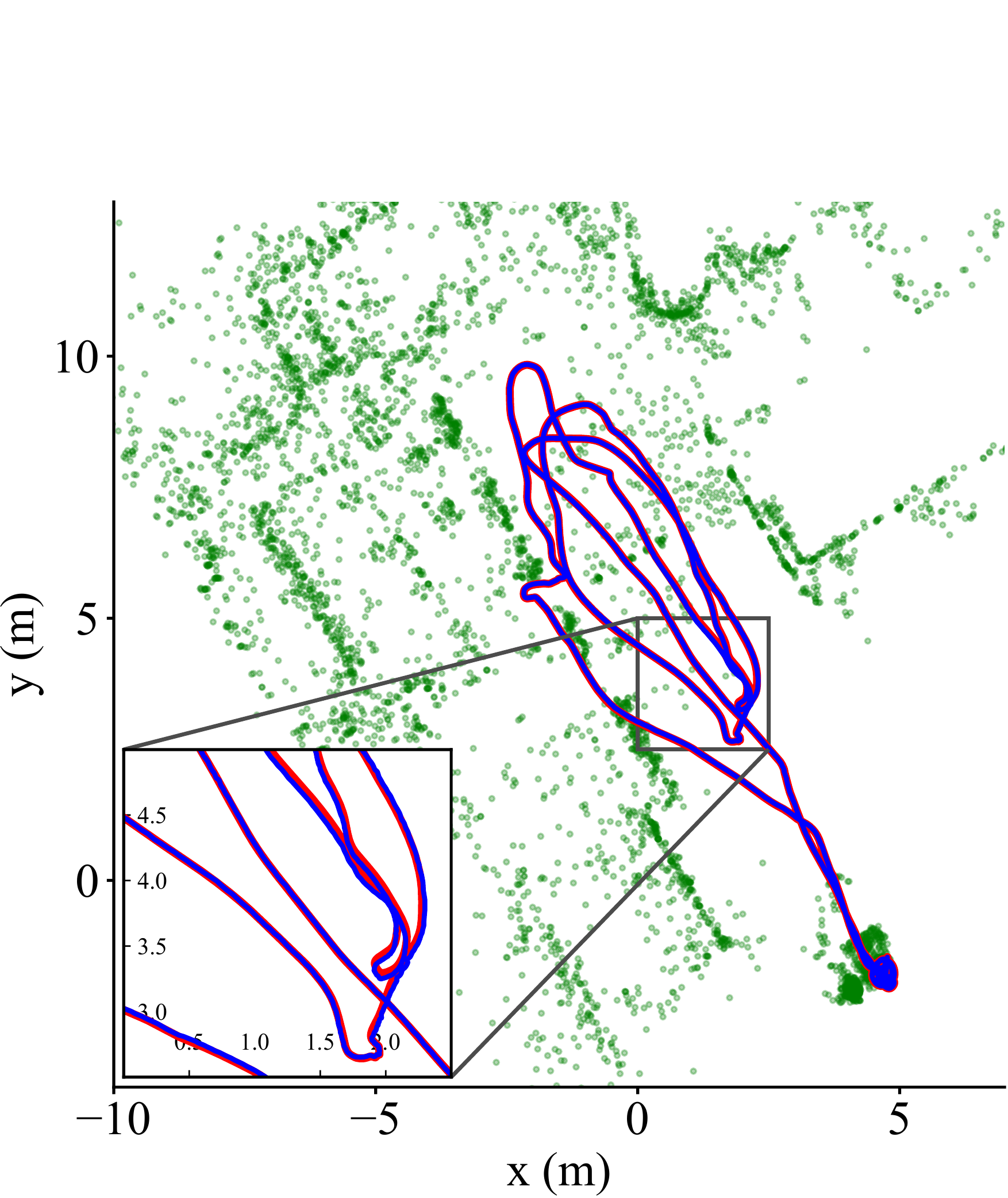}} \hspace*{0.001em}

	\caption{(a) and (b): excerpts of sequnce Machine Hall 01 and 02 the EuRoC dataset, keypoints are colored by the inverse depth. (c) and (d): trajectories of ours against the ground truth on Machine Hall 01 and 02.}
	\label{fig.euroc_qualitative}
\end{figure*}

\begin{table*}[t!]
	\caption{Translational RMSE (cm) on the EuRoC dataset. Left column: performances of pure visual odometry. Right column: performance of SLAM system, where results on sequences without explicit loop-closure are excluded.}
	\label{tab:euroc_acc}
	\centering
	\begin{tabular}{@{}ccccc | cc@{}}
		\toprule
		\multicolumn{2}{c}{\multirow{2}{*}{\textbf{Sequence}}} & \textbf{Ours}   & \multirow{2}{*}{\textbf{DSO}} & \textbf{ORB-SLAM2}           & \multirow{2}{*}{\textbf{Ours}} & \multirow{2}{*}{\textbf{ORB-SLAM2}}                            \\
		                                                       &                 & \textbf{w/o loop}             &                              & \textbf{w/o loop}                                                                               \\ \midrule
		\multirow{6}{*}{easy}                                  & Machine Hall 01 & \textbf{1.63 $\pm$ 0.86}      & 5.73 $\pm$ 2.28              & 1.77     $\pm$   0.91          & -                                   & -                        \\
		                                                       & Machine Hall 02 & \textbf{1.46 $\pm$ 0.77}      & 4.53 $\pm$ 2.71              & 1.72     $\pm$   0.81          & -                                   & -                        \\
		                                                       & Machine Hall 03 & 3.20             $\pm$ 1.53   & 20.89 $\pm$ 10.23            & \textbf{3.19 $\pm$ 1.52}       & -                                   & -                        \\
		                                                       & Vicon Room 1 01 & 3.34             $\pm$ 4.34   & 13.52 $\pm$ 8.72             & \textbf{3.28 $\pm$ 1.20}       & -                                   & -                        \\
		                                                       & Vicon Room 2 01 & \textbf{1.85 $\pm$ 0.83}      & 4.70 $\pm$ 2.41              & 2.59     $\pm$   1.34          & \textbf{1.60     $\pm$   0.15}      & 4.44 $\pm$ 1.31          \\
		                                                       & Vicon Room 2 02 & \textbf{4.73 $\pm$ 2.49}      & 12.31 $\pm$ 6.37             & (5.08   $\pm$   2.95)          & 2.59    $\pm$   0.37                & \textbf{1.69 $\pm$ 0.10} \\ \midrule
		\multirow{5}{*}{hard}                                  & Machine Hall 04 & \textbf{8.59 $\pm$ 3.37}      & 20.03 $\pm$ 8.96             & 9.78    $\pm$   4.05           & -                                   & -                        \\
		                                                       & Machine Hall 05 & \textbf{4.34 $\pm$ 2.09}      & 10.47 $\pm$ 3.78             & (12.67  $\pm$   5.86)          & -                                   & -                        \\
		                                                       & Vicon Room 1 02 & \textbf{23.82 $\pm$ 9.38}     & 32.83 $\pm$ 27.41            & (14.06  $\pm$   5.27 )         & \textbf{1.45 $\pm$ 0.11}            & 1.59 $\pm$ 0.35          \\
		                                                       & Vicon Room 1 03 & (17.17          $\pm$ 10.71)  & \textbf{94.15 $\pm$   39.78} & $\boldsymbol{\times}$          & \textbf{4.44 $\pm$ 1.31}            & 7.26 $\pm$   5.78        \\ \bottomrule
	\end{tabular}
\end{table*}

\subsection{Evaluation on the Local Feature Parameterization}
\label{sec.eval_param}

In this section, we evaluate the effectiveness of different feature parameterization methods in pose estimation.
As the local feature is parameterized as a gaussian distribution, with different combination of mean and variance, we mainly compare four different parameterization methods, given by:
\begin{itemize}
	\item EXTREMA IDENTITY (EI): $\mathbf{u} = \mathbf{u}_\text{peak}, \boldsymbol{\Sigma} = \mathbf{I}_{2\times 2}$
	\item EXTREMA COVARIANCE (EC): $\mathbf{u} = \mathbf{u}_\text{peak}, \boldsymbol{\Sigma} = \boldsymbol{\bar{\Sigma}}$
	\item MEAN IDENTITY (MI): $\mathbf{u} = \mathbf{\bar{u}}, \boldsymbol{\Sigma} = \mathbf{I}_{2\times2}$
	\item MEAN COVARIANCE (MC): $\mathbf{u} = \mathbf{\bar{u}}, \boldsymbol{\Sigma} = \boldsymbol{\bar{\Sigma}}$
\end{itemize}
where the $\mathbf{\bar{u}}$ and $\boldsymbol{\bar{\Sigma}}$ are described in \autoref{sec.tracking}.

For the dataset selection, we notice that the accuracy of SLAM system (or BA) is mainly bounded by the data association, and thus the local feature parameterization is not the most important factor.
Therefore, using a real-world dataset can hardly compare the effectiveness of different parameterization methods.
This is also because of the systematic error in the real-world dataset, for example, the uncertainty of calibration between cameras and sensors providing absolute pose measurements.
As a consequence, here we perform the evaluations on the ICL-NUIM dataset.

To examine how different parameterization methods work in the SLAM system, for each frame in a sequence, we back-project the detected keypoints to the 3D space as landmarks.
Then these landmarks are associated with $N$ (5 in the experiment) adjacent frames.
Combining the poses along with the landmark positions, we jointly optimize these variables following a standard pipeline of BA.
We maintain an optimization problem with identical optimizable parameters and constraints, and introduce different local feature parameterization methods to examine their effectiveness.

\begin{table*}[t!]
	\caption{Translational RMSE (m) on the KITTI Odometry Dataset.}
	\label{tab:kitti_acc}
	\centering
	\begin{tabular}{@{}ccccc | cc@{}}
		\toprule
		\multirow{2}{*}{\textbf{Sequence}} & \multirow{2}{*}{\textbf{Environment}} & \textbf{Ours}                    & \multirow{2}{*}{\textbf{DSO}} & \textbf{ORB-SLAM2}               & \multirow{2}{*}{\textbf{Ours}}  & \multirow{2}{*}{\textbf{ORB-SLAM2}} \\
		                                   &                                       & \textbf{w/o loop}                &                               & \textbf{w/o loop}                                                                                        \\ \midrule
		00                                 & Urban                                 & 96.23     $\pm$   44.76          & 118.58 $\pm$ 0.51             & \textbf{73.39    $\pm$   32.66}  & \textbf{5.31     $\pm$   0.57}  & 6.11     $\pm$   0.32               \\
		01                                 & Highway                               & $\times$                         & \textbf{105.14	$\pm$ 204.86 }  & $\times$                         & -                               & -                                   \\
		02                                 & Urban+Country                         & 76.94   $\pm$   50.34            & 122.67	$\pm$ 6.93              & \textbf{27.14     $\pm$   14.04} & \textbf{22.54     $\pm$   1.46} & 28.92     $\pm$   4.26              \\
		03                                 & Country                               & 1.34     $\pm$   0.29            & 2.08	$\pm$ 0.08                & \textbf{0.96     $\pm$   0.53}   & -                               & -                                   \\
		04                                 & Country                               & \textbf{0.21     $\pm$   0.08}   & 0.99	$\pm$ 0.01                & 1.14     $\pm$   0.58            & \textbf{0.51     $\pm$   0.18}  & 0.87     $\pm$   0.31               \\
		05                                 & Urban                                 & 48.74     $\pm$   26.33          & 50.93	$\pm$ 0.69               & \textbf{39.42     $\pm$   21.05} & \textbf{4.75     $\pm$   0.61}  & 5.65     $\pm$   1.04               \\
		06                                 & Urban                                 & \textbf{54.37     $\pm$   25.51} & 65.81	$\pm$ 1.29               & 54.45     $\pm$  25.21           & \textbf{10.16     $\pm$   1.74} & 16.13     $\pm$   0.83              \\
		07                                 & Urban                                 & 16.96     $\pm$   9.37           & 18.37	$\pm$ 0.54               & \textbf{16.71     $\pm$   9.47}  & 4.16     $\pm$   0.63           & \textbf{2.45     $\pm$   0.31}      \\
		08                                 & Urban+Country                         & 64.82     $\pm$   39.95          & 119.12	$\pm$ 12.74             & \textbf{51.55     $\pm$   31.78} & -                               & -                                   \\
		09                                 & Urban+Country                         & 60.82     $\pm$   31.60          & 61.21	$\pm$ 7.72               &
		\textbf{47.31     $\pm$   25.96}   & \textbf{9.52     $\pm$   0.47}        & 51.63     $\pm$   12.70                                                                                                                                                     \\
		10                                 & Urban+Country                         & \textbf{8.27     $\pm$   0.86}   & 34.84	$\pm$ 38.41              & 9.56     $\pm$   1.02            & -                               & -                                   \\
		\bottomrule
	\end{tabular}
\end{table*}

The results are represented in \autoref{fig.eval_param}, from which we have the following conclusions:
First, the results show that considering the prediction with an anisotropic covariance assigning to different keypoints could generally increase the bundle adjustment accuracy.
For methods that do not consider the uncertainty (EI and MI), the covariance matrices are all set to be identity.
This means that in the optimization, different residual factors contribute to the final results equally.
On the contrary, for methods taking uncertainty into account (EC and MC), the anisotropic covariance serve as a reflection of keypoint prediction uncertainty.
Through modelling the prediction uncertainty, different factors are weighed in the joint optimization according to their observation confidence.
In other words, factors with higher uncertainty should have less contribution to the optimization framework.
As a consequence, comparing EC and MC against EI and MI, the accuracy of BA is increased with modelling the uncertainty.
Second, comparing methods that consider keypoint location as extrema (EI and EC) or weighted mean (MI and MC), we find that the pose accuracy is improved slightly.
This indicates that most of the keypoint predictions are already of good quality, so that in a BA framework, considering the refinement of keypoint location with prediction uncertainty can only provide trivial improvement in the pose estimation.
Accordingly, these experimental results guide us to finally choose the local feature parameterization in our system described in \autoref{sec.tracking}.

\subsection{Evaluation on Trajectory Estimation}

In this section, we evaluate our system on several public datasets, the New Tsukuba \cite{martull2012realistic}, the EuRoC Mav dataset \cite{Burri25012016} and KITTI Odometry dataset \cite{geiger2012we}.
We compare our method to both the state-of-the-art indirect (ORB-SLAM2 \cite{mur2017orb}) and direct (DSO \cite{engel2018dso}) VO/VSLAM algorithms.
GCN-SLAM \cite{tang2019gcnv2}, the recently proposed method using a learned binary descriptor, is expected to be one of the competitors. However, the monocular version adapted from the open-source implementation\footnote{\href{https://github.com/jiexiong2016/GCNv2_SLAM}{https://github.com/jiexiong2016/GCNv2\_SLAM}} fails to produce competitive results.
Thus we exclude it for further evaluation.
A major reason for the failure of GCN-SLAM is that it is designed for RGB-D inputs, while for the monocular VO, maintaining the scale consistency and estimating the depth of local feature raise more challenges.

All the experiments are done using the same desktop with i7-8700K and NVIDIA 1080Ti.
For quantitative evaluation, translational RMSE of absolute trajectory error (ATE) \cite{sturm12iros} is used.
We run different algorithms on each sequence 5 times and average the evaluation metrics.
The tracking failure is either reported by the system itself or determined afterward if the error is larger than 1 meter (typically caused by scale inconsistency).
If tracking failure exists for one or more runs, the corresponding result is shown in parentheses $(\cdot)$, while failure for all runs is marked as $\boldsymbol{\times}$.
The results on both datasets are reported in \autoref{tab:tsukuba_acc}, \autoref{tab:euroc_acc} and \autoref{tab:kitti_acc}, respectively.

\subsubsection{The New Tsukuba Dataset}

The New Tsukuba dataset provides synthetic images rendered by computer graphics techniques.
It is challenging for monocular visual odometry as 1) the illumination condition of sequences \textit{lamps} and \textit{flashlight} is extreme, and 2) the camera rotation is relatively aggressive.
As ORB-SLAM2 does not provide the setting for the Tsukuba dataset, we use the setting of another indoor dataset with the same image resolution and adjust the camera parameters accordingly.
Note that the New Tsukuba Dataset does not contain any loopy motion, therefore all methods can be considered as pure visual odometry rather than full SLAM systems.

\autoref{tab:tsukuba_acc} reports the translational RMSE and success rate of tracking for each sequence.
As expected, challenging illumination conditions of the \textit{lamps} and \textit{flashlight} lead both ORB-SLAM and DSO to tracking failure.
Besides, to accelerate feature association for camera pose tracking, ORB-SLAM searches correspondences in a local window with a motion prior, making it sensitive to aggressive rotation change.
As a consequence, it occasionally fails even under the \textit{fluorescent} with a moderate illumination configuration.

Besides the robustness and accuracy under these test scenarios,
our system is capable of maintaining a relatively consistent trajectory estimation accuracy, regardless of the illumination variances.
Especially on \textit{lamps}, the RMSE is even slightly smaller than on \textit{daylight}.
Larger error under \textit{flashlight} over other sequences indicates that learning-based descriptors share a similar characteristic with hand-crafted ones in degradation under photometric noise \cite{engel2018dso}. 

\subsubsection{EuRoC Mav Dataset}
The EuRoC dataset contains several sequences that are challenging for monocular ego-motion estimation.
Accordingly, for the comparison, we divide the dataset into \textit{moderate} (easy) and \textit{challenging} (hard) sequences.
For the sequences MH01, MH02, MH03, V101, V201 and V202, the camera motion is slow, and the illumination conditions are moderate, making them relatively easy for monocular VO/VSLAM.
On the contrary, in MH04, MH05, and V103, the illumination varies in a wide range.
In V102, V103, and V203, the camera motion is aggressive, including nearly pure rotation or fast movement.
Additionally, for that currently a pinhole camera model is assumed by our system, we pre-rectify all the images in the evaluation, which limits the field-of-view (FOV) of the camera and brings more challenges for monocular VO/VSLAM, especially indirect methods.


Here we report performances for both VO/VSLAM systems.
Due to all the methods fail on V203, the corresponding results are not included.
For the pure VO, as shown in \autoref{tab:euroc_acc}, for the moderate sequences, the proposed system has comparable accuracy with the state-of-the-art methods. 
For the challenging sequences, our method increases the robustness and accuracy compared to the baselines.
Especially for V103, ORB-SLAM is unable to track the camera motion with severe exposure change and aggressive motion continuously, while our method only fails in 1/10 runs, indicating a more robust performance.
Considering the full SLAM system, the proposed system successfully increase the robustness and accuracy compared to the VO-only solution.
It is noticeable that for the difficult sequence V103, we achieve an RMSE of 4.44 cm, which significantly improves the result from the pure VO system.
Compared our method against ORB-SLAM2, we observe our system generally provides a comparable performance with improvement on V103.

\subsubsection{KITTI Odometry Dataset}
The KITTI Odometry Dataset \cite{geiger2012we} is commonly used for evaluating visual state estimation systems in large-scale outdoor scenario.
The result comparisons are reported in \autoref{tab:kitti_acc} for both pure odometry and SLAM.
As in \autoref{tab:kitti_acc}, the pure visual odometry results do not outperform ORB-SLAM2 while the full SLAM results is more competitive, which validate the improvement on introducing loop closure with NetVLAD.
Our system takes more time in data association due to the usage of floating descriptors, which limits the computation time for batched optimization.
However, as more representative local features are selected, our system manages a more lightweight map and advances in global BA with loop closure.
We notice that for the sequences with loops (e.g., 00, 02, 05-07, 09, 10), generally the proposed system performs better than ORB-SLAM2.
Especially for sequence 09, we notice that ORB-SLAM2 failed to detect loop-closure with BoW, leading the trajectory to be inconsistent globally.
On the contrary, our system successfully detects the loop with learnt global embedding. Along with global BA to optimize the trajectory, not surprisingly we achieve a better estimation result.
Exception occurs on 08, where our system exhibits more severe scale drift in trajectory estimation, resulting in larger trajectory error.
For the sequence 01, both methods fail to maintain a correct scale estimation, while only DSO succeeds.

\begin{figure}[t!]
	\centering
	\subfloat[]{\centering\includegraphics[width=0.24\textwidth]{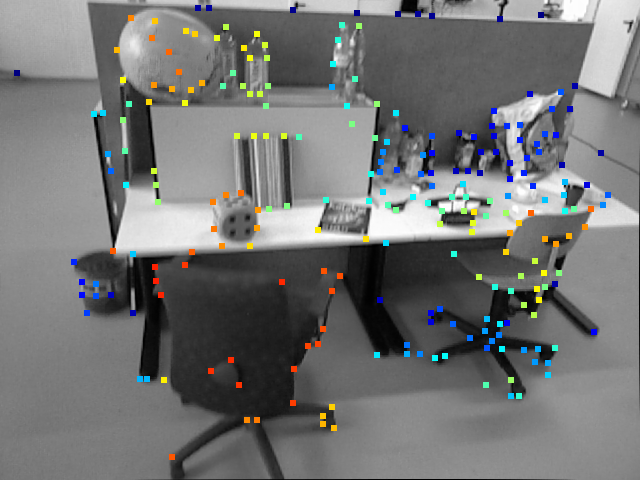}}\enspace
	\subfloat[]{\includegraphics[width=0.24\textwidth]{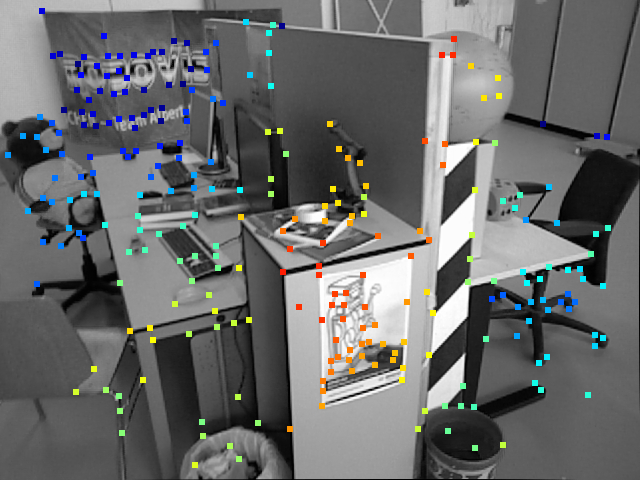}}
	\caption{Excerpts on the TUM Dataset, keypoints are colored by the inverse depth.}
	\label{fig.tum_excerpts1}
\end{figure}
\begin{figure}[t!]
	\centering
	\subfloat[]{\centering\includegraphics[width=0.24\textwidth]{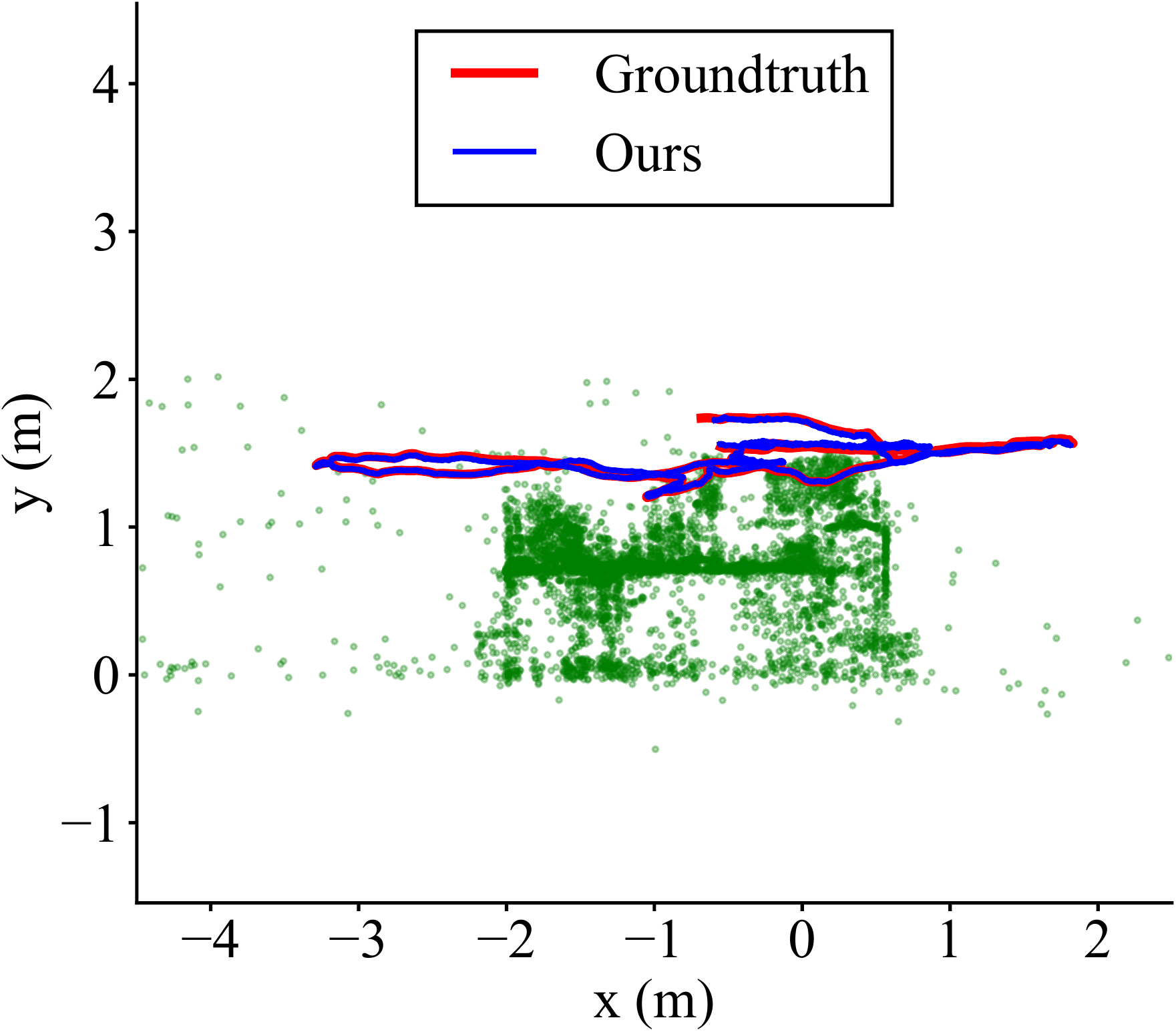}}\enspace
	\subfloat[]{\includegraphics[width=0.24\textwidth]{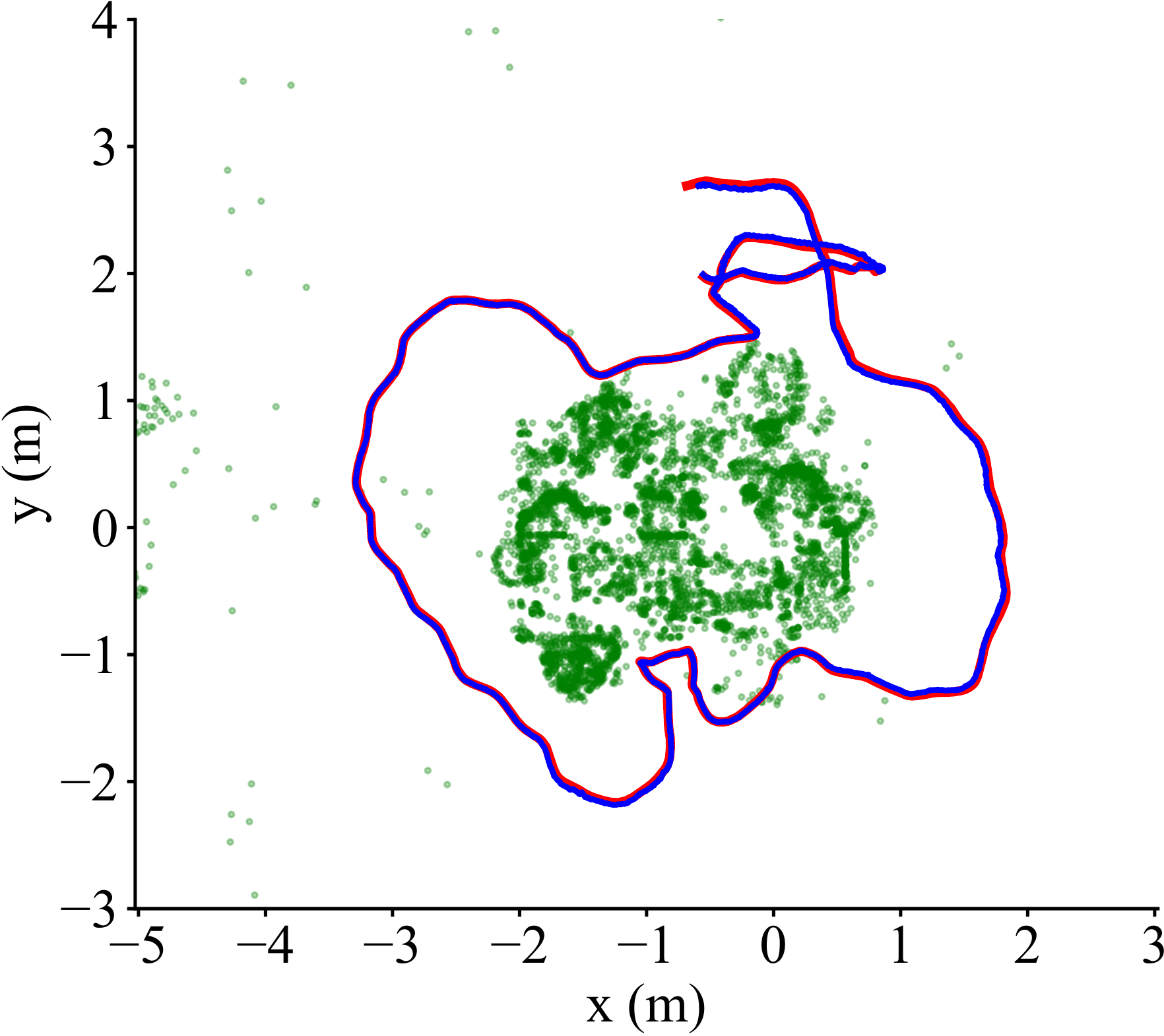}}

	\caption{Qualitative results on TUM Dataset.}
	\label{fig.tum_excerpts2}
\end{figure}

\subsection{Qualitative Results}
Here we also provide some qualitative results in \autoref{fig.euroc_qualitative}, \autoref{fig.tum_excerpts1} and \autoref{fig.tum_excerpts2}.
They show the reconstructed sparse map and estimated trajectories on several sequences.
In \autoref{fig.euroc_qualitative}, even some repetitive features can be well associated and triangulated with relatively accurate depth, for example, features on the floor.
As in \autoref{fig.tum_excerpts1}, in this indoor environment, we notice that our estimated trajectory has only trivial drift without using loop-closure.
In addition, the sparse map well recovers the geometry in the test desk scene.
\autoref{fig.tum_excerpts2} shows that the trajectory along with recovered visual structure on TUM Dataset, where we notice the estimated trajectory is well aligned with the ground truth.

\subsection{Evaluations on Local Reconstruction}

\begin{figure}[t!]
	\centering
	\includegraphics[width=0.36\textwidth]{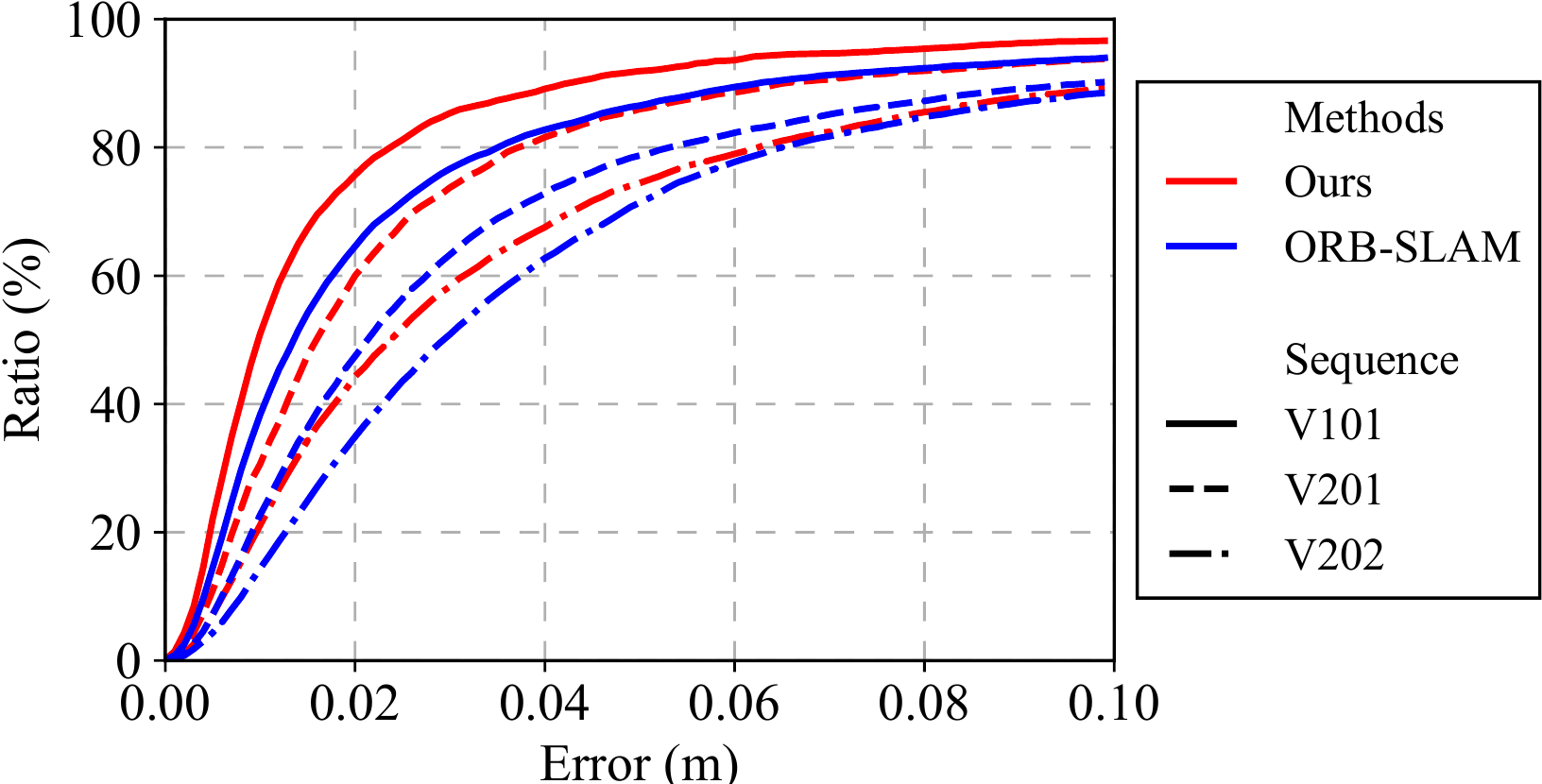}
	\caption{Evaluation on the local reconstructions. The higher the percentage of points towards a zero distance the better.}
	\label{fig.eval_map}
\end{figure}

To evaluate the reconstruction accuracy,
the sparse pointclouds generated by different methods are scaled and transformed via the alignment results.
The error metric is defined as the RMSE of distances to the nearest neighborhood.
The similar evaluation process can be found in \cite{millane2018c}.
Three sequences V101, V201, and V202 of EuRoC that provide dense ground-truth structure are selected for the evaluation.

As shown in \autoref{fig.eval_map}, our method recovers a more accurate local structure, which in turn guarantees the accuracy of local trajectory estimation.
Additionally, in V101, although the ATE of our method does not outperform ORB-SLAM2, the local structure is more accurate than ORB-SLAM2, indicating fewer outliers in our system.
One advantage of indirect methods over direct ones is the covisibility connections established by powerful descriptors, with which local map is fully reused and the VO drift can be reduced \cite{engel2018dso}.
ORB-SLAM2 does a great job of associating features cross views and fusing redundant points.
However, under the cases of wider baseline, our system better associates landmarks with local features.
Especially, comparing the sparse reconstruction of certain objects in the scene (e.g. the checkboard), drastically fewer outliers and redundancies are generated from our system.

\section{Conclusions and Future Work}
\label{sec.conclusions}

In this paper, we presented a monocular SLAM system leveraging learnt description, repeatability and global embedding.
Different from previous work, we focused on tightly coupling the learning-based frontend with indirect multiview geometry to fully exploit the network predictions.
By interpreting the repeatability prediction in a probabilistic manner, w proposed a two-step tracking scheme to estimate the camera pose: direct tracking on the repeatability maps and refining the pose with the patch-wise association.
We further discuss the local feature parameterization to consider the uncertainty in the network prediction.
With loop closure detection from deep global embeddings, we maintain the consistency via global bundle adjustment.
The experimental results demonstrated that the proposed system is capable to handle challenging situations, where both the state-of-the-art indirect and direct methods suffer from strong degradation.

In the future, we would like to investigate more lightweight network architectures or binary descriptor learning to accelerate our system.
In the current system implementation, descriptor matching causes certain overhead for the map management, which we believe can be further optimized.
In addition, supervising the learning frontend in an end-to-end manner might help our system better exploit from data-driven approaches.
For example, using learnt relative pose as a prior for camera pose tracking could boost the general robustness under challenging conditions, in which cases direct formulations could not work well.
Last but not least, extending the current system into a stereo or visual-inertial system is considered to be of much more practical value.
As they can provide absolute scale information, such extension could better help robotic navigation in the long-term operations.





\bibliography{slam,slamdl,slamhdr}
\bibliographystyle{IEEEtran}

\addtolength{\textheight}{-12cm}   

\end{document}